%% file: main.tex
\documentclass[10pt,twocolumn,letterpaper]{article}

\usepackage[pagenumbers]{cvpr}
\input{preamble}

\definecolor{cvprblue}{rgb}{0.21,0.49,0.74}
\usepackage[pagebackref,breaklinks,colorlinks,allcolors=cvprblue]{hyperref}

\title{Back to Raw Motion: Latent-Free Diffusion Transformers\\for Text-to-Motion and Exact Conditional Generation}

\author{Yun Chen \footnotemark[1]\\
[0.3em]
KAIST\\
{\tt\small cyruby@kaist.ac.kr}
\and
Jeonghyeok Do \footnotemark[1]\\
[0.3em]
KAIST\\
{\tt\small ehwjdgur0913@kaist.ac.kr}
\and
Munchurl Kim \footnotemark[2]\\
[0.3em]
KAIST\\
{\tt\small mkimee@kaist.ac.kr}
\and
\small{\url{https://kaist-viclab.github.io/RawMo_site}}
}
\begin{document}
\maketitle

{
  \renewcommand{\thefootnote}{\fnsymbol{footnote}}
  \footnotetext[1]{These authors contributed equally.}
  \footnotetext[2]{Corresponding author.}
}

\input{sec/0_abstract}    
\input{sec/1_intro}

\input{sec/2_related}
\input{sec/3_preliminary}
\input{sec/4_method}
\input{sec/5_experiment}
\input{sec/6_conclusion}

\clearpage

\input{sec/X_suppl}

\clearpage

{
    \small
    \bibliographystyle{ieeenat_fullname}
    \bibliography{main}
}

\end{document}

%% file: preamble.tex
\usepackage{url}
\usepackage{graphicx}

\usepackage{booktabs}       
\usepackage{amsfonts}       
\usepackage{nicefrac}       
\usepackage{xcolor}         

\usepackage{amsmath}
\usepackage{amssymb}
\usepackage{graphicx}
\usepackage{multirow}
\usepackage{bm}
\usepackage{array}
\usepackage{makecell}
\usepackage{adjustbox}
\usepackage[table]{xcolor}
\usepackage{pifont}
\usepackage{algorithm}
\usepackage{algpseudocode}
\usepackage{wrapfig}
\usepackage{cuted}
\usepackage{capt-of}

\newcommand{\sd}[1]{{\tiny$\pm$#1}}

\definecolor{oursbg}{RGB}{253,236,231}
\definecolor{refgray}{gray}{0.55}
\newcommand{\grouprow}[2]{\multicolumn{#1}{l}{\textit{#2}}\\}
\newcommand{\ul}[1]{\underline{#1}}

%% file: sec/0_abstract.tex
\begin{abstract}

As generative models tackle larger and more complex datasets, learned codecs enable efficient generation through compact latent representations, but may lose details and cannot guarantee exact preservation of given conditions. Generating directly in the original data space avoids errors introduced by the codec. Recent advances in image generation demonstrate that pixel diffusion can produce high-quality samples at practical computational costs. Inspired by these advances, we conduct a preliminary study of pixel diffusion, demonstrating its potential for adaptation to motion generation. Building on these findings, we present \textbf{RawMo}, a framework that brings text-to-motion generation to raw motion space. Our Motion Diffusion Transformer (MiT) models local skeletal structures and whole-body dynamics, while refining motion across frames to improve temporal consistency. Working directly in raw motion space further allows our high-noise text-motion alignment loss (HiNCE) to improve text consistency without intermediate motion decoding. RawMo supports multiple motion-conditioning tasks with a single conditional model while preserving given motion conditions exactly. Experiments on HumanML3D and KIT-ML show the best text alignment among compared methods and competitive motion quality, with ablations confirming the benefits of MiT and HiNCE.

\end{abstract}

%% file: sec/1_intro.tex
\section{Introduction}
\label{sec:intro}

\begin{figure}[t]
    \centering
    \includegraphics[width=\linewidth]{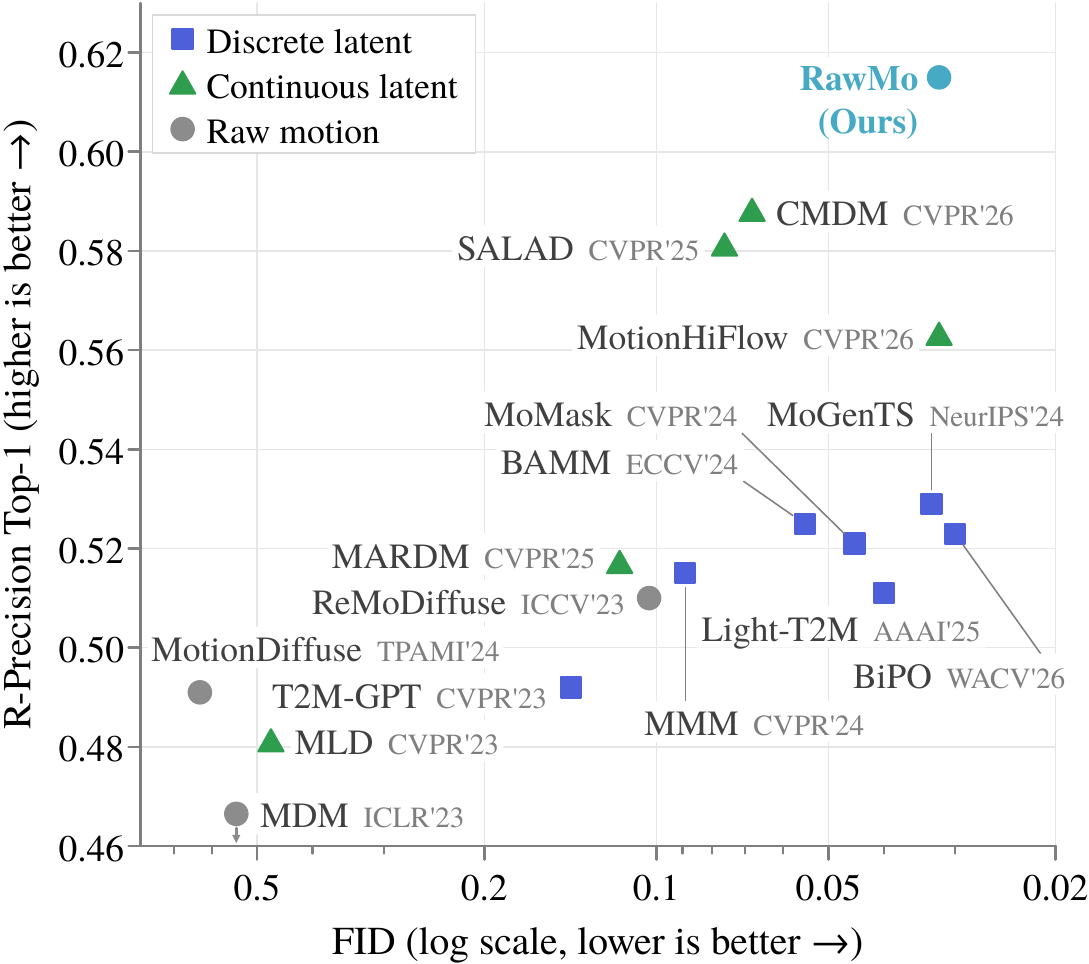}
    \vspace{-0.6cm}
    \caption{\textbf{Text-to-motion generation on HumanML3D.} RawMo achieves the highest R@1 with competitive FID under the \mbox{263-D} evaluator of Guo \etal~\cite{guo2022humanml3d}. Marker shapes distinguish discrete latent, continuous latent, and raw-motion methods. The downward arrow indicates MDM's off-scale R@1 of 0.320.}
    \vspace{-0.3cm}
    \label{fig:teaser}
\end{figure}

Text-to-motion generation produces 3D human motion sequences from natural-language descriptions. The task has broad applications in character animation, games, and virtual environments, where natural and diverse movements are essential. Paired text-motion datasets~\cite{guo2022humanml3d,plappert2016kitml} have supported progress in learning the relationship between language and movement. Early diffusion models generate motion through iterative denoising of full motion sequences~\cite{tevet2023mdm}, but repeated processing can incur substantial sampling costs~\cite{chen2023mld}. To reduce the computational burden, many approaches move generation into compact learned representations, using either continuous latent spaces or discrete motion codes~\cite{chen2023mld,zhang2023t2mgpt,guo2024momask}. Generation in these compact spaces has advanced efficiency while achieving improvements in motion quality and text consistency.

\begin{figure}[t]
    \centering
    \includegraphics[width=\linewidth]{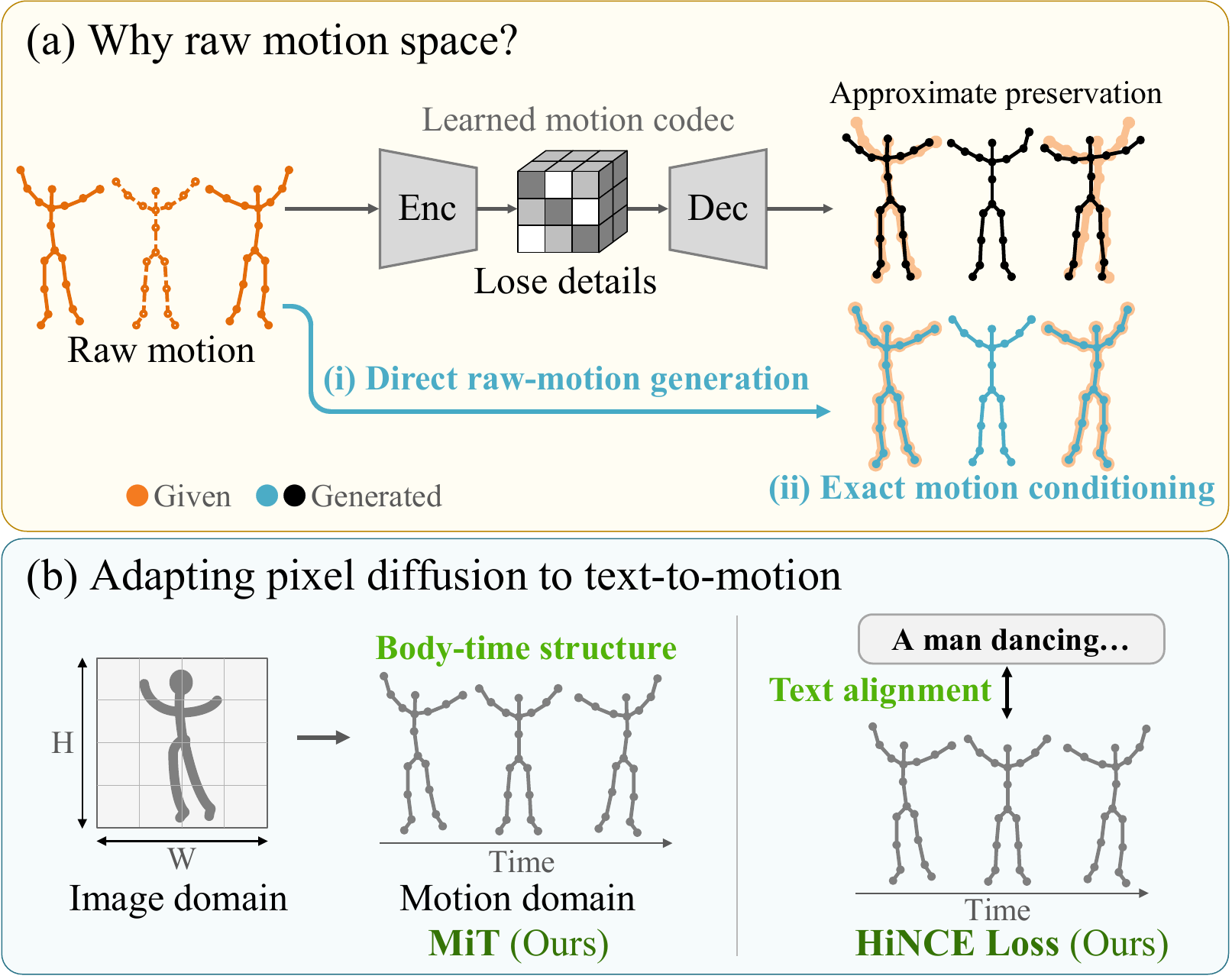}
    \vspace{-0.5cm}
    \caption{\textbf{Motivation for RawMo.} (a) Direct raw-motion generation avoids errors introduced by a learned motion codec and enables exact preservation of given motion conditions. (b) Adapting pixel diffusion to text-to-motion requires modeling body-time structure and aligning generated motion with textual descriptions.}
    \vspace{-0.3cm}
    \label{fig:motivation}
\end{figure}

As generative models tackle larger and more complex data, learned latent spaces offer compact representations that reduce computation. An encoder compresses data into latent codes, and a decoder converts generated latent codes back into data samples. However, details can be lost during encoding and decoding, and latent-space generation alone is unable to guarantee exact preservation of given conditions. Recent work on pixel diffusion in the image domain demonstrates that direct data-space generation can achieve high quality at practical computational costs~\cite{li2026jit,yu2026pixeldit}. We explore this direction in the motion domain, where body regions and frame order can provide the basis for organizing local motion features and modeling temporal changes. Our goal is to use these known body and time relationships to model motion directly, instead of learning a separate compressed representation. Fig.~\ref{fig:motivation}(a) illustrates our motivation for returning to raw motion: (i) avoiding errors introduced by a learned motion codec through direct generation and (ii) preserving given motion conditions exactly rather than matching them only approximately.

JiT~\cite{li2026jit} highlights that a denoising network can predict in one space while its loss is computed in another through output conversion. Inspired by this idea, we conduct a preliminary study comparing prediction targets and loss spaces using a plain JiT-style Transformer over raw motion patches. Consistent with JiT's image experiments, clean-motion prediction with a flow-velocity loss gives the lowest FID among the tested combinations, supporting the feasibility of extending pixel-diffusion principles to motion generation. Section~\ref{sec:preliminary} presents the complete study.

Building on these findings, we bring text-to-motion generation \emph{back to raw motion} with \textbf{RawMo}. Adapting pixel diffusion to motion requires modeling body-part relationships and coordinated movement. Our Motion Diffusion Transformer (MiT) extends PixelDiT's dual-level design~\cite{yu2026pixeldit} to model skeletal structures and whole-body dynamics, with a refinement head that promotes temporal continuity. To strengthen text alignment, we pretrain motion and text encoders on paired training data to establish a shared embedding space for measuring text-motion consistency. During RawMo training, our high-noise text-motion alignment loss (HiNCE) uses these frozen pretrained encoders to supervise clean-motion predictions at high-noise timesteps, guiding the model toward the described action in the early denoising steps. Fig.~\ref{fig:motivation}(b) illustrates how MiT and HiNCE address motion structure and text alignment.

RawMo supports multiple motion-conditioning tasks with a single model, preserving given motion conditions \textit{exactly} while generating the remaining motion. Comparisons on HumanML3D~\cite{guo2022humanml3d} and KIT-ML~\cite{plappert2016kitml} assess motion quality, text consistency, and motion-conditioning performance. Fig.~\ref{fig:teaser} shows HumanML3D results. RawMo achieves the strongest text alignment among compared methods and zero control error in motion-conditioned generation; ablations reveal how motion structure and supervision timing shape the balance between motion quality and text consistency.

Our main contributions are summarized as follows:
\begin{itemize}
    \item We conduct a study demonstrating the feasibility of adapting pixel-diffusion principles to motion generation; the study identifies clean-motion prediction with a flow-velocity objective as an effective combination.
    \item We propose RawMo, which brings text-to-motion generation back to raw motion; MiT models body and time structure, while HiNCE provides semantic supervision at high-noise timesteps.
    \item Experiments on HumanML3D and KIT-ML demonstrate RawMo's superior generation performance, while a single RawMo model supports multiple motion-conditioning tasks with exact preservation of given motion conditions.
\end{itemize}

%% file: sec/2_related.tex
\section{Related Work}
\label{sec:related}

\subsection{Pixel-space image diffusion}

Pixel-space diffusion generates images directly without a learned codec. Early diffusion models performed iterative denoising in pixel space, but processing high-resolution images incurred substantial computational costs~\cite{ho2020ddpm,dhariwal2021adm}. Latent diffusion reduced these costs by moving denoising into compressed representations, introducing dependence on a learned encoder and decoder~\cite{rombach2022ldm}. Recent work revisits pixel-space generation through suitable prediction targets, loss spaces, and efficient architectures~\cite{li2026jit,yu2026pixeldit,chen2026dip}. JiT~\cite{li2026jit} separates prediction targets from loss spaces, showing that clean-image prediction with a velocity-space objective enables plain Transformers to model high-dimensional pixel patches. Architectural developments further separate global modeling from local detail synthesis~\cite{yu2026pixeldit,chen2026dip}. PixelDiT~\cite{yu2026pixeldit} combines patch-level semantic modeling with Pixel Transformer (PiT) blocks for fine-grained pixel updates, while complementary designs use lightweight local refinement~\cite{chen2026dip}. These developments revisit the prediction task and computational structure of direct generation, motivating raw-motion generation that accounts for body structure and temporal continuity.

\subsection{Text-to-motion generation}

Text-to-motion models generate human motion sequences that follow textual descriptions. Early diffusion-based methods denoise motion sequences directly in motion space~\cite{tevet2023mdm,zhang2024motiondiffuse,zhang2023remodiffuse}, with clean-motion prediction enabling geometric supervision on the generated sequence~\cite{tevet2023mdm}. To reduce generation costs, many approaches adopt compact learned motion representations. Continuous methods perform diffusion or flow matching in an autoencoder's latent space~\cite{chen2023mld,li2026motionhiflow,yu2026cmdm}, while discrete methods combine vector quantization with autoregressive or masked token prediction~\cite{zhang2023t2mgpt,guo2024momask,pinyoanuntapong2024bamm,yuan2024mogents}. Skeleton-aware latent designs further capture relationships among joints, frames, and words~\cite{hong2025salad}. Recent work also explores flow matching directly in motion space, with representation-aware noise scaling~\cite{yu2026motionspaceflow}. Our RawMo revisits direct motion generation through body-region modeling, frame-level refinement, and high-noise text-motion supervision informed by recent pixel diffusion.

\subsection{Motion-conditional generation}

Motion-conditional generation synthesizes motion under partial observations, such as joint trajectories, keyframes, or selected body parts. Existing approaches use inference-time inpainting or spatial guidance~\cite{tevet2023mdm,karunratanakul2023gmd}, compose diffusion priors for joint-level control~\cite{shafir2024priormdm}, or use learned control branches~\cite{xie2024omnicontrol,dai2024motionlcm}. Other methods refine structured motion latents through trajectory-based optimization~\cite{wan2024tlcontrol}, while observation-conditioned denoising learns completion from randomly sampled frame and joint observations~\cite{cohan2024condmdi}. Unified frameworks further combine generation and editing through a common formulation of source motion, conditions, and target motion~\cite{guo2025motionlab}. Prior work incorporates partial motion observations during denoising~\cite{tevet2023mdm} and uses sampling-time projection to enforce exact linear spatial constraints~\cite{watanabe2026projflow}. Our RawMo preserves given motion conditions exactly using a single conditional model with a global skip.

%% file: sec/3_preliminary.tex
\section{Preliminary: Back to Raw Motion}
\label{sec:preliminary}

To assess the feasibility of transferring pixel-space diffusion principles to motion, we conduct a preliminary study in raw motion space. JiT~\cite{li2026jit} motivates our study by showing that a plain Transformer can generate high-quality images directly from raw patches without a learned codec.

For a motion sequence $\mathbf{x}\in\mathbb{R}^{T\times D}$ with $T$ frames and $D$ features per frame, we form patches of $s=4$ consecutive frames, yielding $\mathbf{x}_{\mathrm{patch}}\in\mathbb{R}^{(T/s)\times(sD)}$. Our plain baseline applies the same grouping to the noisy sequence, linearly embeds each patch into a Transformer token, and uses a linear output head to reconstruct $s\times D$ motion features per token. The study includes both the plain baseline and RawMo (Sec.~\ref{sec:rawmo}) to examine whether JiT's findings for images hold across motion architectures.

\begin{table}[t]
    \centering
    \caption{\textbf{Prediction targets and loss spaces on HumanML3D.} We report FID$\downarrow$ averaged over three seeds. Rows specify prediction targets and columns specify loss spaces. Both architectures achieve their lowest FID with clean-motion prediction and a velocity-space loss.}
    \vspace{-0.2cm}
    \label{tab:prediction_loss}
    \setlength{\tabcolsep}{7pt}
    \renewcommand{\arraystretch}{1.1}
    \resizebox{\linewidth}{!}{%
    \begin{tabular}{ccccccc}
        \toprule
        \multirow{2}{*}{\textbf{Prediction}}
        & \multicolumn{3}{c}{\textbf{Plain JiT-style}}
        & \multicolumn{3}{c}{\textbf{RawMo (Ours)}} \\
        \cmidrule(lr){2-4}
        \cmidrule(l){5-7}
        & $\mathbf{x}$-loss
        & $\bm{\epsilon}$-loss
        & $\mathbf{v}$-loss
        & $\mathbf{x}$-loss
        & $\bm{\epsilon}$-loss
        & $\mathbf{v}$-loss \\
        \midrule
        $\mathbf{x}$
        & .0450 & .0819
        & \cellcolor{BlueGreen!10}.0350
        & .0328 & .0909
        & \cellcolor{BlueGreen!10}.0317 \\
        $\bm{\epsilon}$
        & 61.68 & 62.95 & 61.77
        & .0477 & .2244 & .0406 \\
        $\mathbf{v}$
        & 9.08 & 8.69 & 8.41
        & .0492 & .0812 & .0355 \\
        \bottomrule
    \end{tabular}%
    }
    \vspace{-0.3cm}
\end{table}

A denoising network does not need to use the same space for prediction and loss computation. Following JiT~\cite{li2026jit}, we denote the network by $f_\theta$ and distinguish the \emph{prediction target} from the \emph{loss space}. The prediction target specifies the direct network output, while the loss space specifies where the converted output is compared with the corresponding ground truth to compute the training loss.

We construct the noisy motion sequence $\mathbf{x}_\tau$ by linearly interpolating between clean motion $\mathbf{x}$ and Gaussian noise $\bm{\epsilon}$ as follows:
\begin{equation}
\mathbf{x}_\tau=\tau\mathbf{x}+(1-\tau)\bm{\epsilon}, \qquad \bm{\epsilon}\sim\mathcal{N}(0,\mathbf{I}),
\label{eq:prelim_path}
\end{equation}
where $\tau=0$ corresponds to pure Gaussian noise and $\tau=1$ to clean motion. The flow velocity along this path is given by $\mathbf{v}=\mathbf{x}-\bm{\epsilon}$. The three prediction targets are linked by the interpolation path: given $\mathbf{x}_\tau$ and $0<\tau<1$, any one of $\mathbf{x}$, $\bm{\epsilon}$, and $\mathbf{v}$ determines the other two. 
For example, a network conditioned on caption $\mathbf{y}$ can output a clean-motion prediction $\hat{\mathbf{x}}=f_\theta(\mathbf{x}_\tau,\tau;\mathbf{y})$. The corresponding noise prediction $\hat{\bm{\epsilon}}$ and velocity prediction $\hat{\mathbf{v}}$ are obtained as:
\begin{equation}
\hat{\bm{\epsilon}}=\frac{\mathbf{x}_\tau-\tau\hat{\mathbf{x}}}{{1-\tau}}, \quad \hat{\mathbf{v}}=\frac{\hat{\mathbf{x}}-\mathbf{x}_\tau}{{1-\tau}}.
\label{eq:prelim_conversion}
\end{equation}
The converted predictions allow a network that directly predicts clean motion to receive supervision in clean-motion, noise, or velocity space. The prediction target and loss space can therefore be chosen separately.

Algebraic conversion does not imply identical training behavior. Although the three prediction targets are convertible, their training formulations differ in network parameterization and timestep-dependent loss weighting. Table~\ref{tab:prediction_loss} evaluates all nine combinations using squared $\ell_2$ losses. Rows indicate the direct network output, and columns indicate the loss space.

\begin{figure*}[t]
\centering
\includegraphics[width=\linewidth]{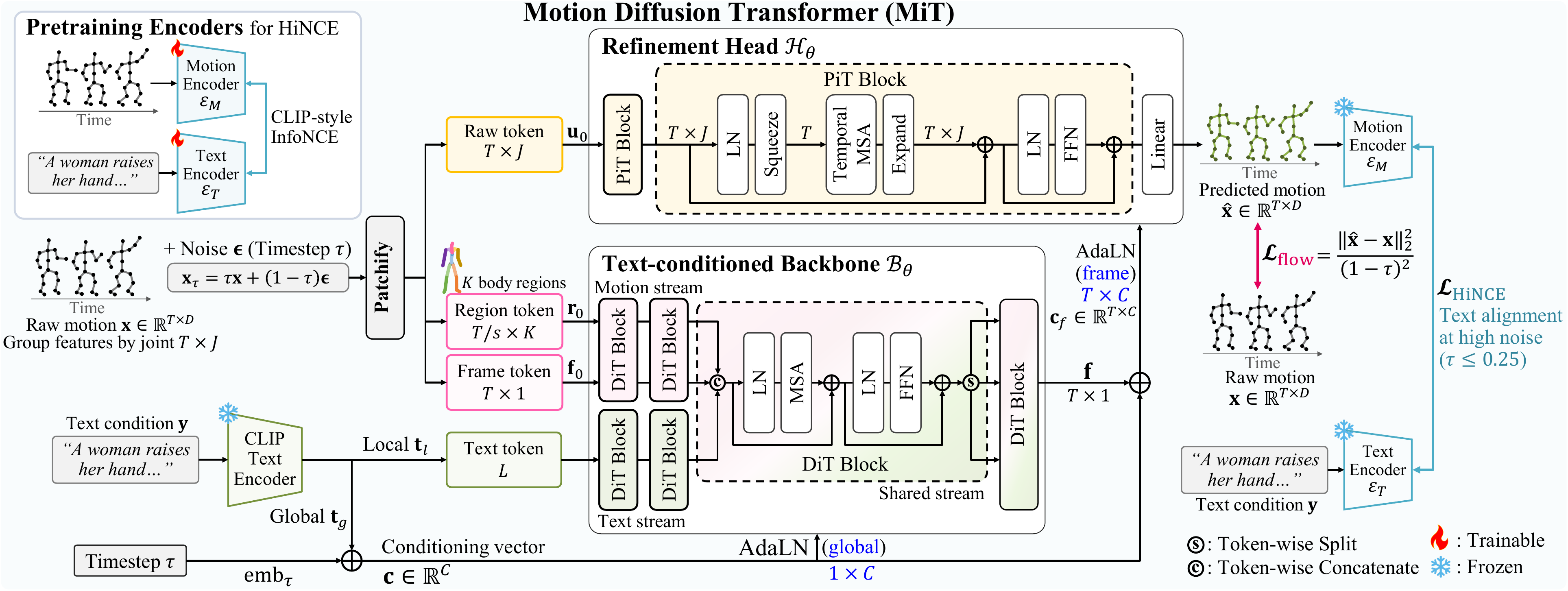}
\vspace{-0.6cm}
\caption{\textbf{RawMo framework.} The Motion Diffusion Transformer (MiT) predicts clean motion through a text-conditioned backbone and a refinement head. The backbone integrates body-region and frame representations with text, while the head refines raw tokens through frame-wise conditioning and temporal attention. Training combines a velocity-space objective with HiNCE, which provides text-alignment supervision at high noise. The motion and text encoders $\mathcal{E}_M$ and $\mathcal{E}_T$ are pretrained for HiNCE and remain frozen during MiT training.}
\vspace{-0.3cm}
\label{fig:framework}
\end{figure*}

The results reproduce JiT's findings for images: both the plain baseline and RawMo achieve their lowest FID with clean-motion prediction and a velocity-space loss. We therefore adopt this combination for RawMo, with the objective defined as:
\begin{equation}
\mathcal{L}_{\mathrm{flow}}=\mathbb{E}\!\left[\|\hat{\mathbf{v}}-\mathbf{v}\|_2^2\right]=\mathbb{E}\!\left[\frac{\|\hat{\mathbf{x}}-\mathbf{x}\|_2^2}{(1-\tau)^2}\right].
\label{eq:flow_loss}
\end{equation}
The agreement across the two architectures supports the feasibility of bringing text-to-motion generation \emph{back to raw motion} through these pixel-diffusion principles and provides an empirical basis for RawMo.

%% file: sec/4_method.tex
\section{RawMo}
\label{sec:rawmo}

Building on the findings in Sec.~\ref{sec:preliminary}, RawMo generates text-conditioned motion directly in raw motion space using clean-motion prediction and a velocity-space objective. Fig.~\ref{fig:framework} provides an overview of RawMo. The Motion Diffusion Transformer (MiT) combines body-region and frame representations in a text-conditioned backbone, followed by joint-level reconstruction through a refinement head (Sec.~\ref{sec:mit}). HiNCE provides text-alignment supervision for clean-motion predictions at high-noise timesteps (Sec.~\ref{sec:hince}). A global skip preserves given motion conditions exactly in each clean-motion prediction (Sec.~\ref{sec:conditional}).

\subsection{Motion Diffusion Transformer (MiT)}
\label{sec:mit}

MiT implements the denoising network $f_\theta$ through a text-conditioned backbone $\mathcal{B}_\theta$ and a refinement head $\mathcal{H}_\theta$, where $\theta$ denotes the trainable parameters. For a motion sequence $\mathbf{x}\in\mathbb{R}^{T\times D}$ with $T$ frames and $D$ motion features per frame ($D=263$ on HumanML3D; $251$ on KIT-ML)~\cite{guo2022humanml3d}, we construct the noisy input $\mathbf{x}_\tau$ at timestep $\tau$ using Eq.~\ref{eq:prelim_path}. The network receives a caption $\mathbf{y}$ as the text condition.

MiT constructs three token types from $\mathbf{x}_\tau$: body-region tokens, frame tokens, and raw tokens. Body-region and frame tokens enter the text-conditioned backbone $\mathcal{B}_\theta$, while raw tokens take a separate path to the refinement head $\mathcal{H}_\theta$.

\noindent\textbf{Text-conditioned backbone.}
Each body-region token represents one of $K=12$ body regions over $s=4$ consecutive frames, providing local skeletal context within a short temporal window. Each frame token represents all $D$ features of a single frame, retaining a sequence of whole-body states for modeling temporal changes. The tokenization operators $\mathcal{P}_{\mathrm{region}}$ and $\mathcal{P}_{\mathrm{frame}}$ group the corresponding motion features and linearly project them into initial region tokens $\mathbf{r}_0$ and frame tokens $\mathbf{f}_0$, respectively, as:
\begin{equation}
\begin{aligned}
\mathbf{r}_0
&=\mathcal{P}_{\mathrm{region}}(\mathbf{x}_\tau)
\in\mathbb{R}^{(T/s)K\times C},\\
\mathbf{f}_0
&=\mathcal{P}_{\mathrm{frame}}(\mathbf{x}_\tau)
\in\mathbb{R}^{T\times C},
\end{aligned}
\label{eq:rawmo_tokens}
\end{equation}
where $C$ denotes the backbone token dimension. Region and frame tokens form the input motion stream $[\mathbf{r}_0;\mathbf{f}_0]$, where $[\cdot;\cdot]$ denotes token concatenation.

A frozen CLIP text encoder~\cite{radford2021clip} encodes the caption $\mathbf{y}$ into a global sentence-level embedding $\mathbf{t}_g\in\mathbb{R}^{C_\mathrm{text}}$ and local token-level embeddings $\mathbf{t}_l\in\mathbb{R}^{L\times C_\mathrm{text}}$, where $C_{\mathrm{text}}$ is the dimension and $L$ is the token count. The global embedding summarizes the entire caption, while the local embeddings provide detailed text context for joint attention with motion. We combine the sentence-level embedding and timestep into the global conditioning vector $\mathbf{c}$ as:
\begin{equation}
\mathbf{c}
=\psi_{\tau}\!\left(\operatorname{emb}_{\tau}\right)
+\psi_g\!\left(\mathbf{t}_g\right)
\in\mathbb{R}^{C},
\label{eq:rawmo_global_condition}
\end{equation}
where $\operatorname{emb}_{\tau}$ is the sinusoidal embedding of timestep $\tau$, and $\psi_{\tau}$ and $\psi_g$ are separate multilayer perceptrons that map their inputs to $C$-dimensional features. The global conditioning vector modulates each backbone block through adaptive normalization, providing timestep and text context.

Following FLUX~\cite{blackforestlabs2024flux}, the backbone processes motion and text through dual-stream blocks followed by single-stream blocks. Dual-stream blocks maintain separate parameters for the motion stream $[\mathbf{r}_0;\mathbf{f}_0]$ and the local text stream $\mathbf{t}_l$, while attention over both streams allows information exchange between motion and text. 
Single-stream blocks then process the concatenated motion and text streams with shared parameters. After the final block, we retain only the updated frame features $\mathbf{f}$, defining the backbone output as:
\begin{equation}
\mathbf{f}
=\mathcal{B}_\theta\!\left(
[\mathbf{r}_0;\mathbf{f}_0;\psi_l(\mathbf{t}_l)];\mathbf{c}\right)
\in\mathbb{R}^{T\times C},
\label{eq:rawmo_backbone}
\end{equation}

\noindent where $\psi_l$ projects local text embeddings to dimension $C$. The output frame features $\mathbf{f}$ incorporate body-region and local text information, providing context for the refinement head $\mathcal{H}_\theta$. The backbone uses rotary positional embeddings (RoPE)~\cite{su2024rope} to encode the temporal position of each motion token. Region tokens use the center of their $s$-frame interval, while frame tokens use their frame indices. 


\noindent\textbf{Refinement head.}
Reconstructing each temporal patch as a separate group of frames can introduce periodic artifacts at patch boundaries. Our refinement head $\mathcal{H}_\theta$ addresses this issue by coordinating joint-level reconstruction across frames through temporal attention. We adapt the Pixel Transformer (PiT) blocks of PixelDiT~\cite{yu2026pixeldit} to motion, grouping joint features by frame. Along the raw-token path, a joint-wise projection $\mathcal{P}_{\mathrm{joint}}$ embeds the noisy motion features of each joint at each frame into raw tokens $\mathbf{u}_0$ as:
\begin{equation}
\mathbf{u}_0
=\mathcal{P}_{\mathrm{joint}}(\mathbf{x}_\tau)
\in\mathbb{R}^{T\times J\times C_h},
\label{eq:rawmo_raw_tokens}
\end{equation}
where $J$ denotes the number of joints and $C_h$ denotes the head token dimension. The backbone frame features $\mathbf{f}$ are combined with the global conditioning vector $\mathbf{c}$ to obtain frame-wise conditioning $\mathbf{c}_f$ as:
\begin{equation}
\mathbf{c}_f
=\mathbf{c}\oplus\mathbf{f}
\in\mathbb{R}^{T\times C},
\label{eq:rawmo_frame_condition}
\end{equation}
where $\oplus$ denotes element-wise addition after repeating $\mathbf{c}$ across the $T$ frames. Each frame's conditioning vector modulates its joint features through adaptive normalization, supplying $\mathcal{H}_\theta$ with timestep, text, and motion context.

Within each head block, a squeeze operation compacts the $J$ joint features of each frame into one vector. Temporal self-attention is then applied across the resulting $T$ frame vectors, which are subsequently expanded back into joint features. A final linear projection maps the refined features back to the original motion channels, producing the clean-motion prediction $\hat{\mathbf{x}}$ as:
\begin{equation}
\begin{aligned}
\hat{\mathbf{x}}
&=f_\theta(\mathbf{x}_\tau,\tau;\mathbf{y}) =\mathcal{H}_\theta(\mathbf{u}_0;\mathbf{c}_f)
\in\mathbb{R}^{T\times D}.
\end{aligned}
\label{eq:rawmo_head}
\end{equation}
Grouping and conditioning operate per frame, while temporal attention coordinates updates across time.


\subsection{Text Alignment at High Noise}
\label{sec:hince}

We use our high-noise text-motion alignment loss (HiNCE) to improve the text consistency of clean-motion predictions. Before MiT training, we jointly pretrain a motion encoder $\mathcal{E}_M$ and a text encoder $\mathcal{E}_T$ on paired training data. Through a CLIP-style InfoNCE objective~\cite{radford2021clip,vandenoord2018infonce}, the two encoders learn a shared embedding space for measuring how well a motion expresses its caption. Fig.~\ref{fig:framework} (top left) illustrates this pretraining stage. 

During MiT training, $\mathcal{E}_M$ and $\mathcal{E}_T$ remain frozen and serve as a fixed semantic reference for HiNCE. Working in raw motion space allows HiNCE to supervise clean-motion predictions directly, passing semantic feedback to MiT without an intermediate motion decoder.

Let $d(\hat{\mathbf{x}},\mathbf{y})$ denote a text-motion distance score between a predicted motion $\hat{\mathbf{x}}$ and caption $\mathbf{y}$, defined as the negative similarity score computed from $\mathcal{E}_M(\hat{\mathbf{x}})$ and $\mathcal{E}_T(\mathbf{y})$. To compute HiNCE over a batch of $N$ motion-caption pairs, we sample a timestep $\tau_i$ for the $i$-th pair. The corresponding clean-motion prediction $\hat{\mathbf{x}}_i$ is paired with caption $\mathbf{y}_i$. We apply alignment supervision at high-noise timesteps to encourage the described action during the early denoising stages. Since smaller $\tau$ corresponds to higher noise, we select the subset $\mathcal{I}=\{i:\tau_i\leq0.25\}$. For a nonempty $\mathcal{I}$, HiNCE is defined as:
\begin{equation}
\mathcal{L}_{\mathrm{HiNCE}}
=-\frac{1}{|\mathcal{I}|}
\sum_{i\in\mathcal{I}}
\log
\frac{\exp[-d(\hat{\mathbf{x}}_i,\mathbf{y}_i)]}
{\sum_{j=1}^{N}\exp[-d(\hat{\mathbf{x}}_i,\mathbf{y}_j)]}.
\label{eq:hince}
\end{equation}
Only high-noise motion predictions contribute to $\mathcal{L}_{\mathrm{HiNCE}}$, while each predicted motion is compared against all $N$ captions, indexed by $j$. We set $\mathcal{L}_{\mathrm{HiNCE}}=0$ when $\mathcal{I}$ is empty.

Combining HiNCE with the velocity-space objective $\mathcal{L}_{\mathrm{flow}}$ from Sec.~\ref{sec:preliminary} gives the overall text-to-motion training objective as:
\begin{equation}
\mathcal{L}
=\mathcal{L}_{\mathrm{flow}}
+\lambda\mathcal{L}_{\mathrm{HiNCE}},
\label{eq:rawmo_objective}
\end{equation}
where $\lambda=0.03$ is the empirically chosen weight of the alignment loss. $\mathcal{E}_M$ and $\mathcal{E}_T$ are used only during training to compute $\mathcal{L}_{\mathrm{HiNCE}}$.

\input{tables/tab_t2m_h3d}
\input{tables/tab_t2m_kit}

\subsection{Exact Motion Conditioning}
\label{sec:conditional}

By generating directly in raw motion space, RawMo can preserve given motion conditions exactly in each clean-motion prediction. Given the noisy motion $\mathbf{x}_\tau$, we train the conditional model with the observed motion $\mathbf{x}_c\in\mathbb{R}^{T\times D}$ and a binary mask $\mathbf{m}\in\{0,1\}^{T\times D}$ as additional input channels. Unobserved entries in $\mathbf{x}_c$ are zero-filled, and $\mathbf{m}$ is $1$ at observed entries and $0$ elsewhere.

A global skip keeps the given motion unchanged and completes the remaining parts using the clean-motion prediction $\hat{\mathbf{x}}$~\cite{tevet2023mdm}. The conditioned clean-motion prediction $\tilde{\mathbf{x}}$ is defined as:
\begin{equation}
\tilde{\mathbf{x}}
=\mathbf{m}\odot\mathbf{x}_c
+(1-\mathbf{m})\odot\hat{\mathbf{x}},
\label{eq:global_skip}
\end{equation}
where $\odot$ denotes element-wise multiplication. The first term carries the given motion directly to the output, while the second supplies predictions only for the unobserved parts. The global skip therefore preserves the given motion conditions exactly in each clean-motion prediction.


%% file: tables/tab_t2m_h3d.tex
\begin{table*}[t]
\scriptsize
\centering
\caption{\textbf{Text-to-motion generation on HumanML3D.} Left: the \mbox{263-D} evaluator~\cite{guo2022humanml3d}; right: the \mbox{67-D} evaluator~\cite{meng2025mardm}. 
RawMo achieves the highest R-Precision and lowest MMDist under both evaluators, with the lowest FID under the 67-D evaluator. Across tables, results are means; $\pm$ denotes standard deviation. \textbf{Bold}/\underline{underline} indicate the best/second-best results; dashes denote unreported results.}
\vspace{-0.2cm}
\label{tab:t2m_h3d}
\setlength{\tabcolsep}{3pt}
\renewcommand{\arraystretch}{1.1}
\resizebox{0.9\textwidth}{!}{%
\begin{tabular}{ll ccccc ccccc}
\toprule
\multicolumn{2}{c}{\textbf{Text-to-motion generation}} & \multicolumn{5}{c}{\textbf{Guo \etal evaluator (263-D)}} & \multicolumn{5}{c}{\textbf{MARDM evaluator (67-D)}}\\
\cmidrule(lr){1-2}\cmidrule(lr){3-7}\cmidrule(l){8-12}
\textbf{Method} & \textbf{Venue} & R@1$\uparrow$ & R@2$\uparrow$ & R@3$\uparrow$ & FID$\downarrow$ & MMDist$\downarrow$ & R@1$\uparrow$ & R@2$\uparrow$ & R@3$\uparrow$ & FID$\downarrow$ & MMDist$\downarrow$\\
\midrule
\rowcolor{gray!10}\multicolumn{12}{l}{\textit{Discrete latent space}}\\
T2M-GPT~\cite{zhang2023t2mgpt} & CVPR'23 & .492 & .679 & .775 & .141 & 3.121 & .470 & .659 & .758 & .335 & 3.505\\
MoMask~\cite{guo2024momask} & CVPR'24 & .521 & .713 & .807 & .045 & 2.958 & .490 & .687 & .786 & .116 & 3.353\\
MMM~\cite{pinyoanuntapong2024mmm} & CVPR'24 & .515 & .708 & .804 & .089 & 2.926 & .487 & .683 & .782 & .132 & 3.359\\
BAMM~\cite{pinyoanuntapong2024bamm} & ECCV'24 & .525 & .720 & .814 & .055 & 2.919 & -- & -- & -- & -- & --\\
MoGenTS~\cite{yuan2024mogents} & NeurIPS'24 & .529 & .719 & .812 & .033 & 2.867 & -- & -- & -- & -- & --\\
Light-T2M~\cite{zeng2025lightt2m} & AAAI'25 & .511 & .699 & .795 & .040 & 3.002 & -- & -- & -- & -- & --\\
BiPO~\cite{hong2026bipo} & WACV'26 & .523 & .714 & .809 & \textbf{.030} & 2.880 & -- & -- & -- & -- & --\\
\rowcolor{gray!10}\multicolumn{12}{l}{\textit{Continuous latent space}}\\
MLD~\cite{chen2023mld} & CVPR'23 & .481 & .673 & .772 & .473 & 3.196 & .461 & .651 & .750 & .431 & 3.445\\
SALAD~\cite{hong2025salad} & CVPR'25 & .581 & .769 & .857 & .076 & 2.649 & .552 & .748 & .839 & .124 & 2.990\\
MARDM~\cite{meng2025mardm} & CVPR'25 & .517 & .708 & .805 & .116 & 2.968 & .500 & .695 & .795 & .114 & 3.270\\
ACMDM~\cite{meng2025acmdm} & arXiv'25 & -- & -- & -- & -- & -- & .522 & .713 & .807 & \ul{.058} & 3.205\\
MotionHiFlow~\cite{li2026motionhiflow} & CVPR'26 & .563 & .754 & .843 & \ul{.032} & 2.691 & -- & -- & -- & -- & --\\
CMDM~\cite{yu2026cmdm} & CVPR'26 & \ul{.588} & \ul{.778} & \ul{.860} & .068 & \ul{2.620} & \ul{.563} & \ul{.759} & \ul{.849} & .078 & \ul{2.920}\\
\rowcolor{gray!10}\multicolumn{12}{l}{\textit{Raw motion space}}\\
MDM~\cite{tevet2023mdm} & ICLR'23 & .320 & .498 & .611 & .544 & 5.566 & .440 & .636 & .742 & .518 & 3.640\\
ReMoDiffuse~\cite{zhang2023remodiffuse} & ICCV'23 & .510 & .698 & .795 & .103 & 2.974 & .468 & .653 & .754 & .883 & 3.414\\
MotionDiffuse~\cite{zhang2024motiondiffuse} & TPAMI'24 & .491 & .681 & .782 & .630 & 3.113 & .450 & .641 & .753 & .778 & 3.490\\
\rowcolor{BlueGreen!10}\textbf{RawMo (Ours)} & -- & \textbf{.615}\sd{.003} & \textbf{.799}\sd{.003} & \textbf{.877}\sd{.001} & \ul{.032}\sd{.006} & \textbf{2.481}\sd{.011} & \textbf{.584}\sd{.002} & \textbf{.774}\sd{.002} & \textbf{.859}\sd{.002} & \textbf{.028}\sd{.006} & \textbf{2.835}\sd{.008}\\
\bottomrule
\end{tabular}}
\vspace{-0.3cm}
\end{table*}

%% file: tables/tab_t2m_kit.tex
\begin{table}[t]
\scriptsize
\centering
\caption{\textbf{Text-to-motion generation on KIT-ML.} Results use the \mbox{251-D} evaluator~\cite{guo2022humanml3d}. RawMo achieves the best scores across all five metrics, covering both motion quality and text alignment.}
\vspace{-0.2cm}
\label{tab:t2m_kit}
\setlength{\tabcolsep}{3pt}
\renewcommand{\arraystretch}{1.1}
\resizebox{\linewidth}{!}{%
\begin{tabular}{lccccc}
\toprule
\textbf{Method} & R@1$\uparrow$ & R@2$\uparrow$ & R@3$\uparrow$ & FID$\downarrow$ & MMDist$\downarrow$ \\
\midrule
\rowcolor{gray!10}\grouprow{6}{Discrete latent space}
T2M-GPT~\cite{zhang2023t2mgpt} & .416 & .627 & .745 & .514 & 3.007 \\
MoMask~\cite{guo2024momask} & .433 & .656 & .781 & .204 & 2.779 \\
BAMM~\cite{pinyoanuntapong2024bamm} & .438 & .661 & .788 & .183 & 2.723 \\
MoGenTS~\cite{yuan2024mogents} & .445 & .671 & .797 & .143 & 2.711 \\
Light-T2M~\cite{zeng2025lightt2m} & .444 & .670 & .794 & .161 & 2.746 \\
\rowcolor{gray!10}\grouprow{6}{Continuous latent space}
MLD~\cite{chen2023mld} & .390 & .609 & .734 & .404 & 3.204 \\
SALAD~\cite{hong2025salad} & .477 & \ul{.711} & \ul{.828} & .296 & 2.585 \\
MoMADiff~\cite{zhang2025momadiff} & .458 & .678 & .797 & \ul{.122} & 2.633 \\
MotionHiFlow~\cite{li2026motionhiflow} & \ul{.482} & .704 & .825 & .135 & \ul{2.552} \\
\rowcolor{gray!10}\grouprow{6}{Raw motion space}
MDM~\cite{tevet2023mdm} & .164 & .291 & .396 & .497 & 9.191 \\
ReMoDiffuse~\cite{zhang2023remodiffuse} & .427 & .641 & .765 & .155 & 2.814 \\
MotionDiffuse~\cite{zhang2024motiondiffuse} & .417 & .621 & .739 & 1.954 & 2.958 \\
\rowcolor{BlueGreen!10}\textbf{RawMo (Ours)} & \textbf{.488}\sd{.011} & \textbf{.716}\sd{.010} & \textbf{.831}\sd{.012} & \textbf{.118}\sd{.013} & \textbf{2.442}\sd{.039} \\
\bottomrule
\end{tabular}}
\vspace{-0.4cm}
\end{table}

%% file: sec/5_experiment.tex
\section{Experiments}
\label{sec:experiment}


\subsection{Datasets and Evaluation Metrics}
\label{sec:datasets_metrics}

\noindent\textbf{Datasets.}
We evaluate RawMo on HumanML3D~\cite{guo2022humanml3d} and KIT Motion-Language (KIT-ML)~\cite{plappert2016kitml}. HumanML3D contains 14,616 motion sequences and 44,970 descriptions (22 joints, 20 fps), while KIT-ML contains 3,911 sequences and 6,278 descriptions (21 joints, 12.5 fps). We follow the standard training, validation, and test splits~\cite{guo2022humanml3d}.

\noindent\textbf{Text-to-motion evaluation.}
We report R-Precision (R@1/2/3) for text-motion retrieval, Fr\'echet Inception Distance (FID) for comparing generated and real motion distributions, and Multimodal Distance (MMDist) for the average embedding distance between paired motions and captions. On HumanML3D, we use both the standard 263-D evaluator of Guo \etal.~\cite{guo2022humanml3d} and the 67-D evaluator introduced by MARDM~\cite{meng2025mardm}. The latter retains root motion and local joint positions. We use both evaluators to verify our performance across different motion representations. KIT-ML uses only the 251-D evaluator of Guo \etal.

\begin{figure*}[t]
\centering

\includegraphics[width=\linewidth]{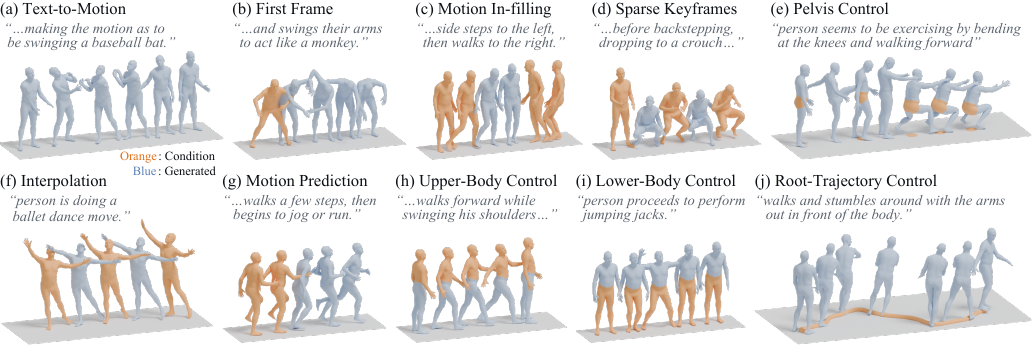}
\vspace{-0.4cm}
\caption{\textbf{Qualitative results of RawMo.} (a)~Text-to-motion generation; (b--j)~motion-conditioned generation using a single conditional model. \textcolor[RGB]{226,113,29}{Orange} indicates given motion conditions, which are preserved exactly; \textcolor[RGB]{91,130,184}{blue} indicates generated motion. The \textit{Appendix} provides visual comparisons for text-to-motion and quantitative and qualitative results for diverse motion-conditioning tasks.}
\label{fig:visual}
\end{figure*}

\input{tables/tab_cond_h3d}

\noindent\textbf{Motion-conditioned evaluation.}
We follow OmniControl~\cite{xie2024omnicontrol} for pelvis control, reporting FID, R@3, Diversity (Div.), and foot skating ratio (Skate). Diversity is better when closer to the real-data reference, while Skate measures foot sliding during ground contact. Trajectory error (Traj.) is the fraction of trajectories with any controlled-frame error above 50\,cm, while location error (Loc.) is the fraction of controlled frames with errors above this threshold. Average error (Err.) is the mean 3-D positional error (m) at controlled frames. We also report FID, R@3, and Div. under the 67-D evaluator.

\subsection{Implementation Details}
\label{sec:exp_implementation}

We train our models from scratch on the training split for 30k steps on HumanML3D and 20k on KIT-ML, using a batch size of 128 on one NVIDIA GeForce RTX 4090 GPU. AdamW uses a learning rate of $2\times10^{-4}$, $(\beta_1,\beta_2)=(0.9,0.95)$, and weight decay $0.01$. The warmup--stable--decay schedule uses 2k linear warmup steps and linear decay to zero over the final 40\% of training. We use bfloat16 and gradient-norm clipping at $1.0$. A frozen CLIP ViT-B/32 encodes captions, with 0.1 caption dropout for classifier-free guidance. HiNCE uses $\lambda=0.03$ for $\tau\leq0.25$. We use a 12-block Transformer backbone (width 384, 6 heads) and a 2-block refinement head. The conditional model follows the HumanML3D settings without HiNCE, taking observations as extra input. Sampling uses Heun's~\cite{karras2022heun} method with 16 intervals (31 network evaluations).

\input{tables/tab_ablation}

\subsection{Text-to-Motion Generation}

\noindent\textbf{Baselines.}
We compare RawMo with methods that generate discrete motion codes~\cite{zhang2023t2mgpt,guo2024momask,pinyoanuntapong2024mmm,pinyoanuntapong2024bamm,yuan2024mogents,zeng2025lightt2m,hong2026bipo}, model continuous motion latents~\cite{chen2023mld,hong2025salad,meng2025mardm,meng2025acmdm,zhang2025momadiff,li2026motionhiflow,yu2026cmdm}, or generate directly in raw motion space~\cite{tevet2023mdm,zhang2023remodiffuse,zhang2024motiondiffuse}.

\noindent\textbf{HumanML3D.}
Table~\ref{tab:t2m_h3d} shows that RawMo achieves the highest R-Precision at all three retrieval levels and the lowest MMDist under both evaluators. With the 263-D evaluator of Guo \etal.~\cite{guo2022humanml3d}, RawMo improves R@1 from 0.588 to 0.615 and reduces MMDist from 2.620 to 2.481 compared with CMDM~\cite{yu2026cmdm}. Although RawMo records a slightly higher FID than BiPO~\cite{hong2026bipo} under the standard evaluator (0.032 versus 0.030), RawMo reaches the highest R@1 of 0.584 and the lowest FID of 0.028 under the 67-D evaluator, compared with an FID of 0.058 for ACMDM~\cite{meng2025acmdm}. The text-alignment advantage holds across both evaluators.

\noindent\textbf{KIT-ML.}
Table~\ref{tab:t2m_kit} shows RawMo leads all five metrics. Compared with MotionHiFlow~\cite{li2026motionhiflow}, RawMo improves R@1 from 0.482 to 0.488 and reduces MMDist from 2.552 to 2.442. RawMo achieves the lowest mean FID (0.118 vs.\ 0.122 for MoMADiff~\cite{zhang2025momadiff}). Results show competitive motion quality and improved text alignment on both datasets.

\subsection{Motion-Conditioned Generation}

\noindent\textbf{Baselines.}
Baselines cover inference-time control using motion priors~\cite{tevet2023mdm,shafir2024priormdm,karunratanakul2023gmd}, learned control mechanisms~\cite{xie2024omnicontrol,dai2024motionlcm}, and optimization or projection for constraint enforcement~\cite{wan2024tlcontrol,watanabe2026projflow}, alongside recent motion-control frameworks~\cite{wang2024intercontrol,pinyoanuntapong2025maskcontrol,meng2025acmdm}.

\noindent\textbf{Pelvis control.}
Pelvis control requires the generated full-body motion to follow pelvis positions specified at selected frames. We preserve height with the global skip and adjust root velocities during sampling to match horizontal targets (see \textit{Appendix}). Table~\ref{tab:cond} reports results on HumanML3D following the protocol of OmniControl~\cite{xie2024omnicontrol}, using a single conditional RawMo model across all observation densities. RawMo achieves \textbf{0 Traj., 0 Loc., and 0 Err.}, demonstrating exact preservation of the specified pelvis positions. Compared with MaskControl~\cite{pinyoanuntapong2025maskcontrol} under the 263-D evaluator of Guo \etal.~\cite{guo2022humanml3d}, RawMo reduces FID from 0.061 to 0.040 and improves R@3 from 0.809 to 0.849. Under the 67-D evaluator~\cite{meng2025mardm}, RawMo also achieves the lowest FID (0.055) and highest R@3 (0.835). These results demonstrate exact spatial control alongside improved motion quality and text alignment under both evaluators. Fig.~\ref{fig:visual} shows qualitative results of our RawMo on text-to-motion generation and diverse motion-conditioning tasks.


\begin{figure}[t]
    \centering
    \includegraphics[width=\linewidth]{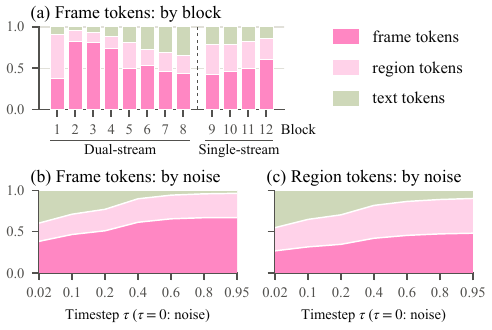}
    \vspace{-0.6cm}
    \caption{\textbf{Sources of attention outputs.} Contributions from frame, region, and text tokens, averaged over five seeds. (a)~By backbone block, averaged over timesteps; (b, c)~by timestep $\tau$, averaged over blocks. Smaller $\tau$ means higher noise.}
    \label{fig:analysis}
\end{figure}

\subsection{Ablations and Attention Analysis}
\label{sec:ablation}

Table~\ref{tab:ablation} examines motion modeling, HiNCE settings, and network sizes on HumanML3D. Fig.~\ref{fig:analysis} shows the relative contributions of frame, region, and text tokens to attention outputs in the trained backbone. The contribution measure is defined in the \textit{Appendix}.

\noindent\textbf{Motion modeling.} 
Removing the refinement head worsens $\mathrm{FID}_{67}$ from 0.028 to 0.060 and MMDist from 2.481 to 2.513. These degradations support our refinement head's role in coordinating reconstruction across frames. Removing either body-region or frame tokens also worsens FID under both evaluators. Fig.~\ref{fig:analysis}(a) shows the sources of frame-token attention outputs by block. Early dual-stream blocks mainly use frame and region information, with text contributions increasing in later dual-stream blocks. In the single-stream stage, frame contributions rise again, and the final frame features provide context for the refinement head. These features already carry region and text information from earlier blocks, so lower direct text contributions do not necessarily indicate weaker text influence.

\noindent\textbf{HiNCE.}
Without HiNCE, R@1 drops from 0.615 to 0.583 and MMDist worsens from 2.481 to 2.605. Extending supervision from our high-noise range ($\tau\leq0.25$) to all timesteps ($\tau\leq1.0$) improves R@1 to 0.620 and MMDist to 2.467, but worsens FID from 0.028 to 0.049 under the 67-D evaluator, showing a trade-off between text alignment and motion quality. Fig.~\ref{fig:analysis}(b,c) shows that text accounts for a larger share of the attention outputs of frame and region tokens at high noise, where HiNCE provides semantic supervision. As the noise level decreases, the relative contributions of frame and region tokens increase.

%% file: tables/tab_cond_h3d.tex
\begin{table*}[t]
\scriptsize
\centering
\caption{\textbf{Motion-conditioned generation on HumanML3D.} Pelvis control follows OmniControl~\cite{xie2024omnicontrol}, averaging over 1, 2, or 5 observed frames and 25\% or 100\% frame coverage. Left: the \mbox{263-D} evaluator; right: the \mbox{67-D} evaluator following ProjFlow~\cite{watanabe2026projflow}. Real-data Diversity is 9.503 (263-D) and 10.455 (67-D). Evaluator-independent skating and control metrics are reported only on the left. A single conditional RawMo model achieves zero Traj., Loc., and Err., together with the best FID and R@3 under both evaluators.}
\vspace{-0.2cm}
\label{tab:cond}
\setlength{\tabcolsep}{6pt}
\renewcommand{\arraystretch}{1.1}
\resizebox{0.95\textwidth}{!}{%
\begin{tabular}{ll ccccccc ccc}
\toprule
\multicolumn{2}{c}{\textbf{Pelvis control} (OmniControl)} & \multicolumn{7}{c}{\textbf{Guo \etal evaluator (263-D)}} & \multicolumn{3}{c}{\textbf{MARDM evaluator (67-D)}}\\
\cmidrule(lr){1-2}\cmidrule(lr){3-9}\cmidrule(l){10-12}
\textbf{Method} & \textbf{Venue} & FID$\downarrow$ & R@3$\uparrow$ & Div. & Skate$\downarrow$ & Traj.$\downarrow$ & Loc.$\downarrow$ & Err.$\downarrow$ & FID$\downarrow$ & R@3$\uparrow$ & Div.\\
\midrule
MDM~\cite{tevet2023mdm} & ICLR'23 & .698 & .602 & 9.197 & .1019 & .4022 & .3076 & .5959 & 1.792 & .673 & 9.131\\
PriorMDM~\cite{shafir2024priormdm} & ICLR'24 & .475 & .583 & 9.156 & .0897 & .3457 & .2132 & .4417 & .393 & .707 & 9.847\\
GMD~\cite{karunratanakul2023gmd} & ICCV'23 & .576 & .665 & 9.206 & .1009 & .0931 & .0321 & .1439 & .238 & .763 & 10.011\\
OmniControl~\cite{xie2024omnicontrol} & ICLR'24 & .218 & .687 & 9.422 & \textbf{.0547} & .0387 & .0096 & .0338 & .081 & .789 & 10.323\\
InterControl~\cite{wang2024intercontrol} & NeurIPS'24 & .159 & .671 & 9.482 & .0729 & .0132 & \ul{.0004} & .0496 & -- & -- & --\\
MotionLCM~\cite{dai2024motionlcm} & ECCV'24 & .531 & .752 & 9.253 & -- & .1887 & .0769 & .1897 & -- & -- & --\\
TLControl~\cite{wan2024tlcontrol} & ECCV'24 & .271 & .779 & 9.569 & -- & \textbf{0} & \textbf{0} & .0108 & -- & -- & --\\
MaskControl~\cite{pinyoanuntapong2025maskcontrol} & ICCV'25 & \ul{.061} & \ul{.809} & 9.496 & \textbf{.0547} & \textbf{0} & \textbf{0} & \ul{.0098} & \ul{.066} & .799 & 10.474\\
ACMDM~\cite{meng2025acmdm} & arXiv'25 & -- & -- & -- & \ul{.0591} & \ul{.0075} & .0010 & .0100 & .067 & \ul{.805} & 10.481\\
ProjFlow~\cite{watanabe2026projflow} & CVPR'26 & -- & -- & -- & .0629 & \textbf{0} & \textbf{0} & \textbf{0} & .107 & .784 & 10.644\\
\rowcolor{BlueGreen!10}\textbf{RawMo (Ours)} & -- & \textbf{.040} & \textbf{.849} & 9.685 & .0674 & \textbf{0} & \textbf{0} & \textbf{0} & \textbf{.055} & \textbf{.835} & 10.767\\
\bottomrule
\end{tabular}}
\vspace{-0.3cm}
\end{table*}

%% file: tables/tab_ablation.tex
\begin{table}[t]
\scriptsize
\centering
\caption{\textbf{Ablation studies on HumanML3D.} We examine the refinement head, token types, HiNCE settings, and network size. R@1, FID, and MMDist use the \mbox{263-D} evaluator; $\mathrm{FID}_{67}$ denotes FID under the \mbox{67-D} evaluator.}
\vspace{-0.2cm}
\label{tab:ablation}
\setlength{\tabcolsep}{3pt}
\renewcommand{\arraystretch}{1.1}
\resizebox{\linewidth}{!}{%
\begin{tabular}{lccccc}
\toprule
\textbf{Variant} & \textbf{Params} & R@1$\uparrow$ & FID$\downarrow$ & MMDist$\downarrow$ & FID$_\text{67}$$\downarrow$ \\
\midrule
\rowcolor{gray!10}\grouprow{6}{Refinement head}
w/o $\mathcal{H}_{\theta}$ & 41.7M & .610\sd{.003} & .039\sd{.004} & 2.513\sd{.007} & .060\sd{.019}\\
\rowcolor{gray!10}\grouprow{6}{Token type}
w/o $\mathbf{r}_0$ & 45.4M & .612\sd{.003} & .034\sd{.004} & 2.492\sd{.011} & .036\sd{.006}\\
w/o $\mathbf{f}_0$ & 42.2M & .612\sd{.002} & .044\sd{.006} & 2.496\sd{.012} & .046\sd{.007}\\
\rowcolor{gray!10}\grouprow{6}{Loss function}
w/o HiNCE & 45.7M & .583\sd{.001} & .035\sd{.004} & 2.605\sd{.009} & .029\sd{.003} \\
$\tau \leq 0.1$ & 45.7M & .601\sd{.002} & .039\sd{.006} & 2.531\sd{.004} & .039\sd{.005} \\
$\tau \leq 1.0$ & 45.7M & .620\sd{.001} & .031\sd{.003} & 2.467\sd{.008} & .049\sd{.005} \\
$\lambda$=0.015 & 45.7M & .607\sd{.003} & .036\sd{.004} & 2.515\sd{.007} & .031\sd{.004} \\
$\lambda$=0.06 & 45.7M & .613\sd{.002} & .031\sd{.005} & 2.482\sd{.014} & .029\sd{.006} \\
\rowcolor{gray!10}\grouprow{6}{Network size}
Width 256 & 22.3M & .606\sd{.002} & .028\sd{.001} & 2.514\sd{.011} & .029\sd{.006} \\
Width 512 & 78.0M & .615\sd{.003} & .039\sd{.006} & 2.472\sd{.010} & .031\sd{.002} \\
Depth 16 & 53.4M & .613\sd{.002} & .035\sd{.002} & 2.485\sd{.013} & .031\sd{.005} \\
Depth 24 & 68.8M & .614\sd{.002} & .036\sd{.006} & 2.482\sd{.053} & .030\sd{.002} \\
\midrule
\rowcolor{BlueGreen!10}\textbf{RawMo} (Full) & 45.7M & .615\sd{.003} & .032\sd{.006} & 2.481\sd{.011} & .028\sd{.006} \\

\bottomrule
\end{tabular}}
\end{table}

%% file: sec/6_conclusion.tex
\section{Conclusion}
\label{sec:conclusion}

We presented \textbf{RawMo}, bringing text-to-motion generation \emph{back to raw motion}. MiT captures skeletal structure and temporal dynamics, while HiNCE improves text alignment through clean-motion supervision at high noise. Experiments on HumanML3D and KIT-ML demonstrate the potential of direct raw-motion generation for high-quality, text-consistent motion with exact condition preservation.

%% file: sec/X_suppl.tex
\clearpage
\appendix

\begin{table}[h]
\scriptsize
\centering
\caption{\textbf{Overview of the \textit{Appendix}.}}
\label{tab:supple_overview}
\setlength{\tabcolsep}{5pt}
\renewcommand{\arraystretch}{1.1}
\resizebox{1.0\linewidth}{!}{%
\begin{tabular}{cl}
\toprule
\rowcolor{BlueGreen!10}
\textbf{Section} & \textbf{Contents} \\
\midrule
Sec.~\ref{supp:large_corpora} & Text-to-motion generation on large-scale datasets \\
Sec.~\ref{supp:conditioned_tasks} & Baseline comparisons across nine conditioning tasks \\
Sec.~\ref{supp:kit} & Additional ablations on KIT-ML \\
Sec.~\ref{supp:prediction_loss} & Prediction targets and loss spaces \\
Sec.~\ref{supp:cost} & Computational cost on HumanML3D \\
Sec.~\ref{supp:visual} & Additional visual comparisons \\
Sec.~\ref{supp:implementation} & Implementation details \\
Sec.~\ref{supp:limitations} & Limitations of zero-shot control and sampling latency \\
\bottomrule
\end{tabular}%
}
\end{table}

\noindent This supplement extends the main paper with text-to-motion results on MotionMillion and RoMo, baseline comparisons across nine HumanML3D motion-conditioning tasks, and ablation studies on KIT-ML. Additional experiments compare all nine combinations of prediction target and loss space on HumanML3D and KIT-ML and report computational costs on HumanML3D. We also provide visual comparisons and implementation details, including attention contribution measurement, and discuss limitations and future work in zero-shot motion control and sampling speed. Table~\ref{tab:supple_overview} summarizes the contents.


\section{Text-to-Motion Generation on Large-Scale Datasets}
\label{supp:large_corpora}

We extend the HumanML3D and KIT-ML experiments to two large-scale motion datasets, MotionMillion~\cite{fan2025motionmillion} and RoMo~\cite{zhang2026romo}, following their official splits. On both datasets, RawMo directly generates the provided 272-D motion features at 30\,fps. Each baseline is retrained separately on each dataset using its official implementation and training schedule for each stage. Checkpoints and, where applicable, guidance scales are selected based on the lowest validation FID. 

Table~\ref{tab:supp_t2m_272} reports results on both datasets. On MotionMillion, RawMo ranks first on all five metrics, reducing FID from 19.79 (MotionStreamer~\cite{xiao2025motionstreamer}) to 9.38 and MM-Dist from 23.58 (CMDM~\cite{yu2026cmdm}) to 21.18. On RoMo, RawMo achieves the lowest FID of 8.57, followed by MotionStreamer at 12.92, and ranks second on R@1 (0.531) and MM-Dist (30.02).

\begin{table*}[t]
\centering
\caption{\textbf{Text-to-motion generation on MotionMillion and RoMo.} We extend evaluation to two large-scale motion datasets. For each dataset, we evaluate all methods on a fixed subset of 8,192 clips selected from its own test set. Results are averaged over three random seeds. Both datasets are evaluated with the MotionMillion evaluator, making the RoMo evaluation cross-corpus. RawMo achieves the best results on all five metrics on MotionMillion and the lowest FID on RoMo.}
\label{tab:supp_t2m_272}
\setlength{\tabcolsep}{3pt}
\renewcommand{\arraystretch}{1.1}
\resizebox{0.9\textwidth}{!}{%
\begin{tabular}{llccccccccccc}
\toprule
\multirow{2}{*}{\textbf{Method}} & \multirow{2}{*}{\textbf{Venue}} & \multicolumn{5}{c}{\textbf{MotionMillion}} & \multicolumn{5}{c}{\textbf{RoMo}} \\
\cmidrule(lr){3-7}\cmidrule(l){8-12}
& & R@1$\uparrow$ & R@2$\uparrow$ & R@3$\uparrow$ & FID$\downarrow$ & MM-Dist$\downarrow$ & R@1$\uparrow$ & R@2$\uparrow$ & R@3$\uparrow$ & FID$\downarrow$ & MM-Dist$\downarrow$ \\
\midrule
\rowcolor{gray!10}\multicolumn{12}{l}{\textit{Discrete latent space}} \\
T2M-GPT~\cite{zhang2023t2mgpt} & CVPR'23 & .372 & .497 & .569 & 222.70 & 34.51 & .223 & .350 & .440 & 140.57 & 34.82 \\
MoMask~\cite{guo2024momask} & CVPR'24 & .737 & .881 & .929 & 51.98 & 26.72 & .504 & .675 & .762 & 47.00 & 30.38 \\
MMM~\cite{pinyoanuntapong2024mmm} & CVPR'24 & .591 & .718 & .778 & 112.41 & 29.98 & .305 & .443 & .534 & 82.88 & 33.89 \\
BAMM~\cite{pinyoanuntapong2024bamm} & ECCV'24 & .769 & .905 & .948 & 30.50 & 25.55 & .508 & .672 & .759 & 34.33 & 30.31 \\
MoGenTS~\cite{yuan2024mogents} & NeurIPS'24 & .760 & .896 & .941 & 50.23 & 25.90 & .520 & .683 & .765 & 45.10 & 30.09 \\
Light-T2M~\cite{zeng2025lightt2m} & AAAI'25 & .636 & .793 & .862 & 32.56 & 28.57 & .462 & .627 & .719 & 30.90 & 30.76 \\
MotionMillion~\cite{fan2025motionmillion} & ICCV'25 & .468 & .636 & .721 & 40.98 & 32.33 & .281 & .418 & .517 & 25.56 & 33.75 \\
\rowcolor{gray!10}\multicolumn{12}{l}{\textit{Continuous latent space}} \\
MLD~\cite{chen2023mld} & CVPR'23 & .517 & .692 & .775 & 34.07 & 31.48 & .306 & .456 & .551 & 26.37 & 33.39 \\
SALAD~\cite{hong2025salad} & CVPR'25 & .614 & .772 & .843 & 58.68 & 29.26 & .523 & \underline{.693} & \underline{.780} & 32.56 & 30.21 \\
MARDM~\cite{meng2025mardm} & CVPR'25 & .578 & .759 & .837 & 33.00 & 29.85 & .414 & .580 & .677 & 14.98 & 31.44 \\
MotionStreamer~\cite{xiao2025motionstreamer} & ICCV'25 & .823 & .933 & .964 & \underline{19.79} & 23.92 & .492 & .655 & .745 & \underline{12.92} & 30.54 \\
ACMDM~\cite{meng2025acmdm} & arXiv'25 & .571 & .746 & .825 & 30.53 & 30.11 & .363 & .522 & .618 & 13.47 & 32.44 \\
MotionHiFlow~\cite{li2026motionhiflow} & CVPR'26 & .773 & .901 & .944 & 28.03 & 25.23 & .492 & .656 & .743 & 38.08 & 30.58 \\
CMDM~\cite{yu2026cmdm} & CVPR'26 & \underline{.828} & \underline{.939} & \underline{.970} & 24.14 & \underline{23.58} & \textbf{.569} & \textbf{.735} & \textbf{.813} & 18.42 & \textbf{29.22} \\
\rowcolor{gray!10}\multicolumn{12}{l}{\textit{Raw motion space}} \\
MDM~\cite{tevet2023mdm} & ICLR'23 & .341 & .502 & .599 & 153.56 & 35.67 & .236 & .367 & .463 & 293.53 & 35.71 \\
ReMoDiffuse~\cite{zhang2023remodiffuse} & ICCV'23 & .443 & .609 & .696 & 130.45 & 33.69 & .402 & .559 & .656 & 26.09 & 32.16 \\
MotionDiffuse~\cite{zhang2024motiondiffuse} & TPAMI'24 & .185 & .294 & .367 & 122.75 & 38.51 & .238 & .362 & .449 & 46.46 & 34.90 \\
\rowcolor{BlueGreen!10}\textbf{RawMo} & -- & \textbf{.900}\sd{.001} & \textbf{.970}\sd{.001} & \textbf{.986}\sd{.001} & \textbf{9.38}\sd{0.21} & \textbf{21.18}\sd{0.02} & \underline{.531}\sd{.002} & .691\sd{.002} & .774\sd{.002} & \textbf{8.57}\sd{0.41} & \underline{30.02}\sd{0.04} \\
\bottomrule
\end{tabular}}
\end{table*}


\section{Motion-Conditioned Generation on HumanML3D}
\label{supp:conditioned_tasks}

We extend the main paper's pelvis-control comparison to the other eight motion-conditioning tasks in Fig.~\ref{fig:visual}. OmniControl~\cite{xie2024omnicontrol}, MaskControl~\cite{pinyoanuntapong2025maskcontrol}, and ACMDM~\cite{meng2025acmdm} are fine-tuned from their released control checkpoints on the same mixture of nine tasks using their official training code. These baselines receive the specified joint positions in global coordinates.

Table~\ref{tab:rawmo_conditioned_tasks} shows that RawMo achieves the highest R@3 and zero error at the conditioned joint positions on all nine tasks. RawMo also achieves the lowest FID on eight tasks and ranks second on sparse keyframes, with FID 0.033 compared with 0.024 for ACMDM. Averaged across tasks, RawMo obtains FID 0.033 and R@3 0.837; the second-best values are 0.044 for ACMDM and 0.813 for MaskControl, respectively.

\begin{table*}[t]
\centering
\caption{\textbf{Motion-conditioned generation on HumanML3D.} The nine tasks correspond to Fig.~\ref{fig:visual}(b)--(j) of the main paper. Results are averaged over five training seeds per method, with one test evaluation per seed. FID and R@3 use the 263-D evaluator of Guo \textit{\textit{et al.}}~\cite{guo2022humanml3d}. Err.\ measures the mean 3D positional error (cm) at the specified joints and frames.}
\label{tab:rawmo_conditioned_tasks}
\small
\setlength{\tabcolsep}{6pt}
\renewcommand{\arraystretch}{1.1}
\resizebox{0.9\textwidth}{!}{%
\begin{tabular}{lcccccccccccc}
\toprule
\multirow{2}{*}{\textbf{Task}} & \multicolumn{3}{c}{OmniControl~\cite{xie2024omnicontrol}} & \multicolumn{3}{c}{MaskControl~\cite{pinyoanuntapong2025maskcontrol}} & \multicolumn{3}{c}{ACMDM~\cite{meng2025acmdm}} & \multicolumn{3}{c}{\cellcolor{BlueGreen!10}\textbf{RawMo (Ours)}} \\
\cmidrule(lr){2-4}\cmidrule(lr){5-7}\cmidrule(lr){8-10}\cmidrule(l){11-13}
& FID$\downarrow$ & R@3$\uparrow$ & Err.$\downarrow$ & FID$\downarrow$ & R@3$\uparrow$ & Err.$\downarrow$ & FID$\downarrow$ & R@3$\uparrow$ & Err.$\downarrow$ & FID$\downarrow$ & R@3$\uparrow$ & Err.$\downarrow$ \\
\midrule
(b) First Frame & .175 & .784 & 9.4 & \underline{.065} & .805 & \underline{1.0} & .086 & \underline{.808} & 4.8 & \textbf{.056} & \textbf{.855} & \textbf{0} \\
(c) Motion In-filling & .075 & .796 & 7.8 & .047 & \underline{.821} & \underline{2.6} & \underline{.039} & .802 & 3.9 & \textbf{.028} & \textbf{.831} & \textbf{0} \\
(d) Sparse Keyframes & .184 & .787 & 7.1 & .054 & \underline{.820} & \underline{2.3} & \textbf{.024} & .814 & 4.8 & \underline{.033} & \textbf{.838} & \textbf{0} \\
(e) Pelvis Control & .144 & .791 & 4.2 & .077 & .805 & \underline{1.0} & \underline{.055} & \underline{.810} & 5.6 & \textbf{.040} & \textbf{.849} & \textbf{0} \\
(f) Interpolation & .124 & .804 & 8.1 & \underline{.025} & \underline{.820} & \underline{2.9} & .026 & .806 & 4.4 & \textbf{.020} & \textbf{.821} & \textbf{0} \\
(g) Motion Prediction & .106 & .798 & \underline{1.4} & .068 & \underline{.816} & 1.9 & \underline{.040} & .807 & 3.1 & \textbf{.029} & \textbf{.845} & \textbf{0} \\
(h) Upper-Body Control & .083 & .798 & 6.5 & \underline{.030} & \underline{.806} & \underline{2.7} & \underline{.030} & .798 & 3.5 & \textbf{.029} & \textbf{.822} & \textbf{0} \\
(i) Lower-Body Control & .103 & .798 & 5.2 & \underline{.043} & \underline{.810} & \underline{2.6} & .050 & .805 & 3.4 & \textbf{.021} & \textbf{.819} & \textbf{0} \\
(j) Root-Trajectory Control & .129 & .792 & 5.7 & .052 & \underline{.810} & \underline{1.8} & \underline{.046} & .807 & 3.2 & \textbf{.045} & \textbf{.849} & \textbf{0} \\
\midrule
Average & .125 & .794 & 6.2 & .051 & \underline{.813} & \underline{2.1} & \underline{.044} & .806 & 4.1 & \textbf{.033} & \textbf{.837} & \textbf{0} \\
\bottomrule
\end{tabular}}
\end{table*}


\section{Ablations on KIT-ML}
\label{supp:kit}

Table~\ref{tab:ablation_kit} extends the main paper's HumanML3D ablations to KIT-ML~\cite{plappert2016kitml}. Removing the refinement head $\mathcal{H}_{\theta}$ produces the largest degradation in mean FID, from 0.118 to 0.177. Without frame tokens $\mathbf{f}_0$, FID worsens to 0.150, compared with 0.124 without body-region tokens $\mathbf{r}_0$.

The full KIT-ML model applies HiNCE at $\tau\leq0.25$ with weight $\lambda=0.02$. Removing HiNCE lowers R@1 from 0.488 to 0.477. Applying HiNCE at all timesteps improves R@1 to 0.495 but worsens FID to 0.132. Increasing backbone depth to 16 gives the lowest mean FID (0.115), with 57.5M parameters compared with 49.9M for the full model. Overall, both datasets show similar trends when removing motion-modeling components or changing the HiNCE timestep range.

\begin{table}[t]
\centering
\caption{\textbf{Ablation studies on KIT-ML.} We examine the refinement head, token types, HiNCE settings, and network size using the 251-D evaluator of Guo \textit{et al.}~\cite{guo2022humanml3d}. For each of five training seeds, we select the guidance scale on the validation set and average ten test evaluations. Results are reported as mean $\pm$ standard deviation across the five training seeds.}
\label{tab:ablation_kit}
\small
\setlength{\tabcolsep}{5pt}
\renewcommand{\arraystretch}{1.1}
\resizebox{\linewidth}{!}{%
\begin{tabular}{lccccc}
\toprule
\textbf{Variant} & \textbf{Params} & R@1$\uparrow$ & R@3$\uparrow$ & FID$\downarrow$ & MMDist$\downarrow$ \\
\midrule
\rowcolor{gray!10}\multicolumn{6}{l}{\textit{Refinement head}} \\
w/o $\mathcal{H}_\theta$ & 41.6M & .472\sd{.006} & .822\sd{.003} & .177\sd{.021} & 2.542\sd{.039} \\
\rowcolor{gray!10}\multicolumn{6}{l}{\textit{Token type}} \\
w/o $\mathbf{r}_0$ & 49.5M & .484\sd{.006} & .828\sd{.005} & .124\sd{.005} & 2.464\sd{.021} \\
w/o $\mathbf{f}_0$ & 43.3M & .483\sd{.002} & .826\sd{.005} & .150\sd{.025} & 2.506\sd{.026} \\
\rowcolor{gray!10}\multicolumn{6}{l}{\textit{Loss function}} \\
w/o HiNCE & 49.9M & .477\sd{.006} & .820\sd{.008} & .133\sd{.012} & 2.481\sd{.025} \\
$\tau\leq0.10$ & 49.9M & .481\sd{.008} & .823\sd{.005} & .142\sd{.008} & 2.471\sd{.017} \\
$\tau\leq1$ & 49.9M & .495\sd{.006} & .834\sd{.008} & .132\sd{.036} & 2.446\sd{.040} \\
$\lambda=0.01$ & 49.9M & .478\sd{.006} & .821\sd{.002} & .126\sd{.006} & 2.493\sd{.012} \\
$\lambda=0.04$ & 49.9M & .485\sd{.009} & .823\sd{.003} & .121\sd{.008} & 2.476\sd{.009} \\
\rowcolor{gray!10}\multicolumn{6}{l}{\textit{Network size}} \\
Width 256 & 25.8M & .486\sd{.011} & .825\sd{.003} & .131\sd{.015} & 2.479\sd{.018} \\
Width 512 & 82.8M & .482\sd{.003} & .824\sd{.003} & .119\sd{.010} & 2.480\sd{.023} \\
Depth 16 & 57.5M & .489\sd{.006} & .832\sd{.002} & .115\sd{.004} & 2.451\sd{.017} \\
Depth 24 & 72.9M & .486\sd{.002} & .824\sd{.005} & .116\sd{.008} & 2.460\sd{.011} \\
\midrule
\rowcolor{BlueGreen!10}\textbf{RawMo} (Full) & 49.9M & .488\sd{.011} & .831\sd{.012} & .118\sd{.013} & 2.442\sd{.039} \\
\bottomrule
\end{tabular}}
\end{table}


\section{Prediction Targets and Loss Spaces}
\label{supp:prediction_loss}

Table~\ref{tab:prediction_loss} in the main paper compares prediction targets and loss spaces for a plain JiT-style Transformer~\cite{li2026jit} and RawMo on HumanML3D over three seeds. Table~\ref{tab:supp_prediction_loss} in this supplement extends this comparison to both HumanML3D and KIT-ML, with the architecture fixed to RawMo. The prediction target and loss space are independently chosen from clean motion $\mathbf{x}$, velocity $\mathbf{v}$, and noise $\boldsymbol{\epsilon}$, giving nine combinations. For each combination, we convert the network output into the selected loss space and compute squared error against the corresponding ground truth. RawMo adopts clean-motion prediction with a velocity-space loss, for which the per-sample error is $\|\hat{\mathbf{x}}-\mathbf{x}\|_2^2/(1-\tau)^2$.

Clean-motion prediction with a velocity-space loss gives the highest HumanML3D R@1 of 0.615 and the lowest $\mathrm{FID}_{67}$ of 0.028. Under the 263-D evaluator on HumanML3D, clean-motion prediction gives the same reported FID of 0.032 with either a velocity-space loss or a clean-motion loss. On KIT-ML, clean-motion prediction with a velocity-space loss gives the lowest FID of 0.118. Velocity prediction with a noise-space loss gives the highest KIT-ML R@1 of 0.496, but FID worsens to 0.196.


\section{Computational Cost}
\label{supp:cost}

Table~\ref{tab:cost} compares the computational cost of the methods evaluated on HumanML3D in the main paper. Generator parameter counts exclude the frozen text encoder. Generation time includes text encoding, sampling, and motion decoding for methods that use a decoder. NFE denotes the number of function evaluations during sampling, with a plus sign separating successive sampling stages. Passes counts individual generator forward passes.

MARDM~\cite{meng2025mardm} and ACMDM~\cite{meng2025acmdm} use adaptive solvers, so their evaluation counts can vary. We measure MARDM runtime using weights we trained. ACMDM runtime is estimated assuming an NFE of 134, which is the median evaluation count of its smaller released model. BiPO~\cite{hong2026bipo} runtime is omitted because pretrained weights were unavailable for our measurements.

RawMo uses a 45.7M-parameter generator and 16 Heun sampling intervals~\cite{karras2022heun}, requiring 31 generator evaluations and 62 forward passes with classifier-free guidance. Generation takes 0.594\,s for one clip and 28.1\,ms per clip at batch size 128. Compared with RawMo, ReMoDiffuse~\cite{zhang2023remodiffuse} is faster for single-clip generation at 0.269\,s, but slower at batch size 128 with 42.5\,ms per clip.

\begin{table}[t]
\centering
\caption{\textbf{Prediction targets and loss spaces for RawMo.} Results are averaged over five training seeds, with the guidance scale selected on the validation set for each seed. On HumanML3D, R@1 and FID use the 263-D evaluator of Guo \textit{et al.}~\cite{guo2022humanml3d}, while $\mathrm{FID}_{67}$ uses the 67-D evaluator~\cite{meng2025mardm}. KIT-ML results use the 251-D evaluator of Guo \textit{et al.}}
\label{tab:supp_prediction_loss}
\small
\setlength{\tabcolsep}{6pt}
\renewcommand{\arraystretch}{1.1}
\resizebox{0.9\linewidth}{!}{%
\begin{tabular}{ccccccc}
\toprule
\multirow{2}{*}{\textbf{Predict}} &  \multirow{2}{*}{\textbf{Loss}} & \multicolumn{3}{c}{\textbf{HumanML3D}} & \multicolumn{2}{c}{\textbf{KIT-ML}} \\
\cmidrule(lr){3-5}\cmidrule(l){6-7}
& & R@1$\uparrow$ & FID$\downarrow$ & FID$_{\text{67}}$$\downarrow$ & R@1$\uparrow$ & FID$\downarrow$ \\
\midrule
$\mathbf{x}$ & $\mathbf{x}$ & 0.609 & \textbf{0.032} & 0.040 & 0.485 & \underline{0.123} \\
\rowcolor{BlueGreen!10}$\mathbf{x}$ & $\mathbf{v}$ & \textbf{0.615} & \textbf{0.032} & \textbf{0.028} & 0.488 & \textbf{0.118} \\
$\mathbf{x}$ & $\boldsymbol\epsilon$   & 0.579 & 0.089 & 0.109 & 0.484 & 0.175 \\
\midrule[0.2pt]
$\mathbf{v}$ & $\mathbf{x}$ & 0.604 & 0.050 & 0.105 & 0.481 & 0.216 \\
$\mathbf{v}$ & $\mathbf{v}$ & 0.603 & \underline{0.038} & \underline{0.039} & 0.484 & 0.168 \\
$\mathbf{v}$ & $\boldsymbol\epsilon$   & 0.590 & 0.079 & 0.102 & \textbf{0.496} & 0.196 \\
\midrule[0.2pt]
$\boldsymbol\epsilon$ & $\mathbf{x}$ & 0.608 & 0.051 & 0.059 & 0.470 & 0.437 \\
$\boldsymbol\epsilon$ & $\mathbf{v}$ & \underline{0.611} & 0.041 & 0.044 & 0.478 & 0.168 \\
$\boldsymbol\epsilon$ & $\boldsymbol\epsilon$   & 0.577 & 0.209 & 0.347 & \underline{0.490} & 0.167 \\
\bottomrule
\end{tabular}%
}
\end{table}

\begin{table*}[t]
\centering
\caption{\textbf{Computational cost on HumanML3D.} NFE denotes the number of function evaluations during sampling, counting each classifier-free guided prediction once. Passes counts individual generator forward passes. Both counts use 196-frame clips. Measured times are medians on a single NVIDIA RTX 4090 using each test clip's length, up to 196 frames. ACMDM~\cite{meng2025acmdm} times are estimates.}
\label{tab:cost}
\small
\setlength{\tabcolsep}{4pt}
\renewcommand{\arraystretch}{1.1}
\resizebox{\textwidth}{!}{%
\begin{tabular}{lcccccccc}
\toprule
\multirow{2}{*}{\textbf{Method}} & \multicolumn{2}{c}{\textbf{Autoencoder}} & \multirow{2}{*}{\textbf{Generator params}} & \multirow{2}{*}{\textbf{Text encoder}} & \multirow{2}{*}{\textbf{NFE}} & \multirow{2}{*}{\textbf{Passes}} & \multicolumn{2}{c}{\textbf{Time}} \\
\cmidrule(lr){2-3}\cmidrule(l){8-9}
& Type & Params & & & & & 1 clip (s) & Batch 128 (ms/clip) \\
\midrule
\rowcolor{gray!10}\multicolumn{9}{l}{\textit{Discrete latent space}} \\
T2M-GPT~\cite{zhang2023t2mgpt}      & VQ-VAE          & 19.4M  & 228.4M & CLIP-B     & 50
   & 50               & 0.218          & 186.4 \\
MoMask~\cite{guo2024momask}       & RVQ-VAE         & 19.4M  &  25.4M & CLIP-B     & $10+5$                    & 30               & 0.086          & 2.7 \\
MMM~\cite{pinyoanuntapong2024mmm}          & VQ-VAE          & 18.0M  & 236.3M & CLIP-B     & 10                        & 10               & 0.080          & 6.3 \\
BAMM~\cite{pinyoanuntapong2024bamm}         & RVQ-VAE         & 19.4M  &  25.4M & CLIP-B     & $98+5$                    & 206              & 0.308          & 8.4 \\
MoGenTS~\cite{yuan2024mogents}      & RVQ-VAE         & 113.8M &  99.4M & CLIP-B     & $20+10$                   & 60               & 0.324          & 20.1 \\
Light-T2M~\cite{zeng2025lightt2m}    & --              & --     &   4.5M & CLIP-B     & 10                        & 20               & 0.112          & 1.5 \\
BiPO~\cite{hong2026bipo} & 6 part-specific VQ-VAEs & 6.3M & 166.0M & CLIP-B & $51+25$             & 76               & --             & -- \\
\rowcolor{gray!10}\multicolumn{9}{l}{\textit{Continuous latent space}} \\
MLD~\cite{chen2023mld}          & VAE             & 18.0M  &   8.3M & CLIP-L     & 50                        & 100              & 0.145          & 3.8 \\
SALAD~\cite{hong2025salad}        & VAE             & 0.17M  &   9.9M & CLIP-B     & 50                        & 100              & 0.298          & 23.0 \\
MARDM~\cite{meng2025mardm}        & AE              & 18.8M  & 290.8M & CLIP-B     & $18+\text{adaptive}$ & $36+3.6\text{k}$ & 3.976 & 189.0 \\
ACMDM~\cite{meng2025acmdm}        & AE              & 17.1M  & 596.2M & CLIP-B     & adaptive      & $2\times\text{NFE}$ & $2.218$ & $1334.0$ \\
MotionHiFlow~\cite{li2026motionhiflow} & VAE             & 2.4M   &  30.8M & CLIP-B     & 14                        & 28               & 0.200          & 8.1 \\
CMDM~\cite{yu2026cmdm}         & VAE             & 76.3M  &  38.7M & DistilBERT & 82                        & 164              & 0.928          & 11.7 \\
\rowcolor{gray!10}\multicolumn{9}{l}{\textit{Raw motion space}} \\
MDM~\cite{tevet2023mdm}          & --              & --     &  17.9M & CLIP-B     & 1000                      & 2000             & 4.473          & 399.6 \\
ReMoDiffuse~\cite{zhang2023remodiffuse}  & --              & --     &  47.0M & CLIP-B     & 50                        & 200              & 0.269          & 42.5 \\
MotionDiffuse~\cite{zhang2024motiondiffuse} & --             & --     &  87.2M & CLIP-B     & 1000                      & 1000             & 8.078          & 504.5 \\
\rowcolor{BlueGreen!10}\textbf{RawMo (Ours)} & -- & -- & 45.7M & CLIP-B & 31 & 62 & 0.594 & 28.1 \\
\bottomrule
\end{tabular}}
\end{table*}


\section{Additional Visual Comparisons}
\label{supp:visual}

\begin{figure*}[tp]
    \centering
    \includegraphics[width=\linewidth]{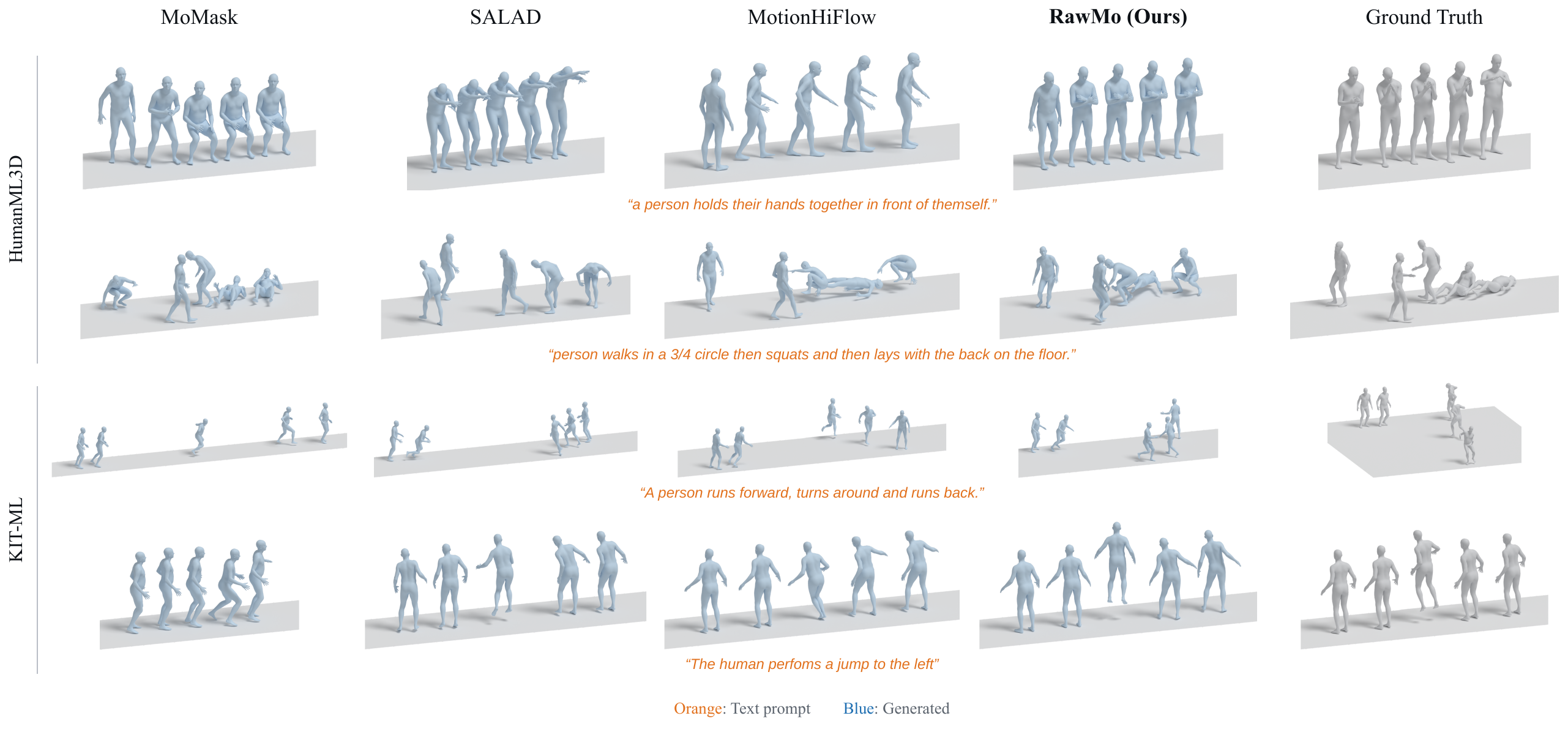}
    \caption{\textbf{Text-to-motion generation on HumanML3D and KIT-ML.} We extend the text-to-motion visualization in Fig.~\ref{fig:visual}(a) of the main paper with comparisons against MoMask, SALAD, and MotionHiFlow. The top two rows show HumanML3D examples and the bottom two show KIT-ML examples. Methods in each row receive the text prompt shown in orange. Across these visual comparisons, generated motion is blue and the corresponding ground-truth reference is gray in the rightmost column.}
    \label{fig:rawmo_visual_t2m}
\end{figure*}

\noindent\textbf{Text-to-motion generation.} Fig.~\ref{fig:rawmo_visual_t2m} extends Fig.~\ref{fig:visual}(a) of the main paper with comparisons against MoMask, SALAD, and MotionHiFlow on HumanML3D and KIT-ML. In the first HumanML3D example, the character generated by RawMo holds both hands together in front of the torso, matching the text prompt. In the KIT-ML example of a jump to the left, RawMo generates an airborne pose.

\begin{figure*}[tp]
    \centering
    \includegraphics[width=\linewidth]{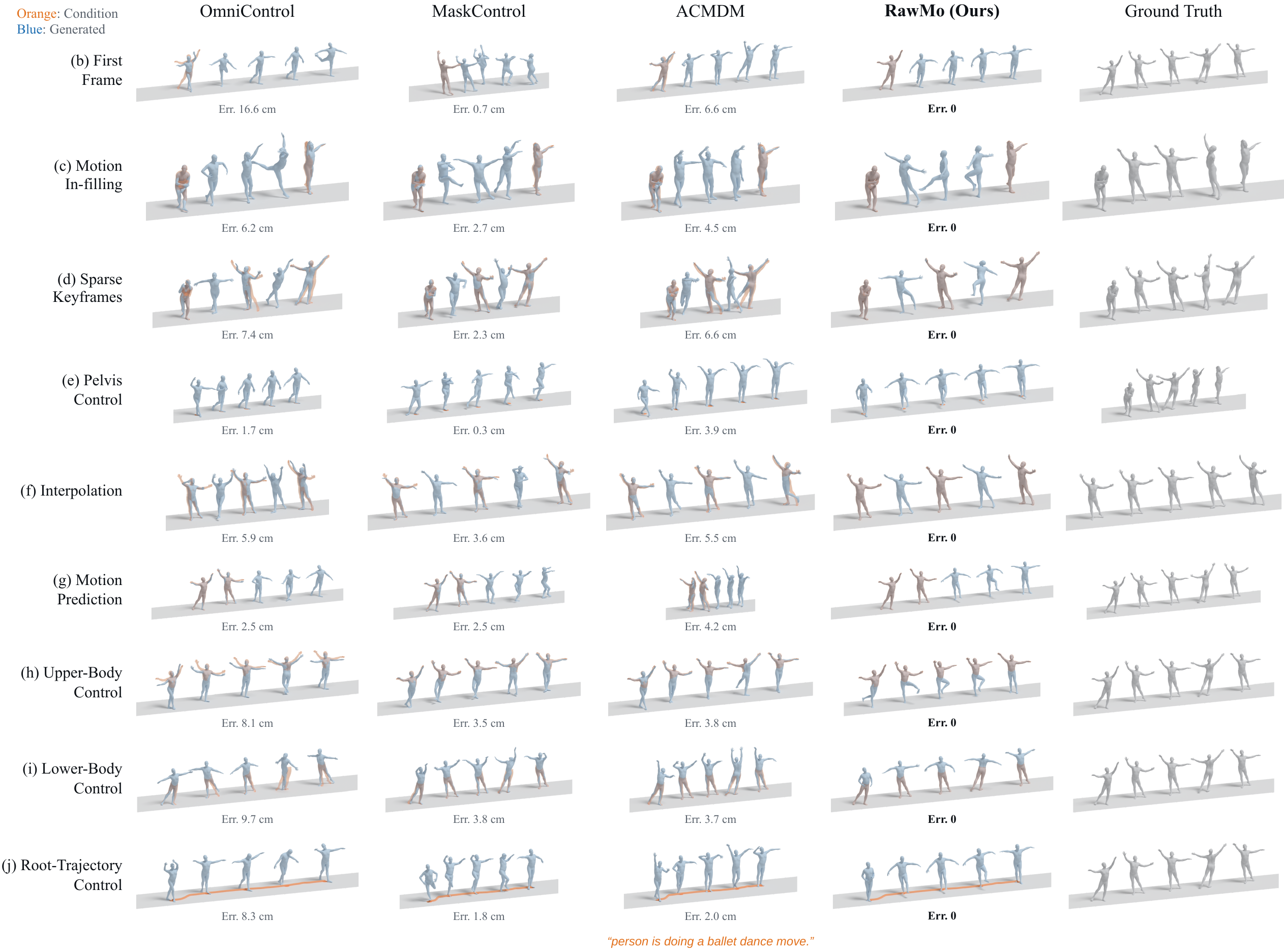}
    \caption{\textbf{Motion-conditioned generation: a ballet example.} We extend Fig.~\ref{fig:visual}(b)--(j) of the main paper by comparing RawMo with OmniControl, MaskControl, and ACMDM on nine motion-conditioning tasks. All rows use one HumanML3D sequence and its text prompt, with different motion conditions. In the visual comparisons, orange marks the given conditions and blue shows generated motion. Err.\ reports the mean 3D positional error in centimeters at the conditioned joint positions for the shown example. A single conditional RawMo model preserves the conditions exactly across all nine tasks.}
    \label{fig:rawmo_visual_ballet}
\end{figure*}

\begin{figure*}[tp]
    \centering
    \includegraphics[width=\linewidth]{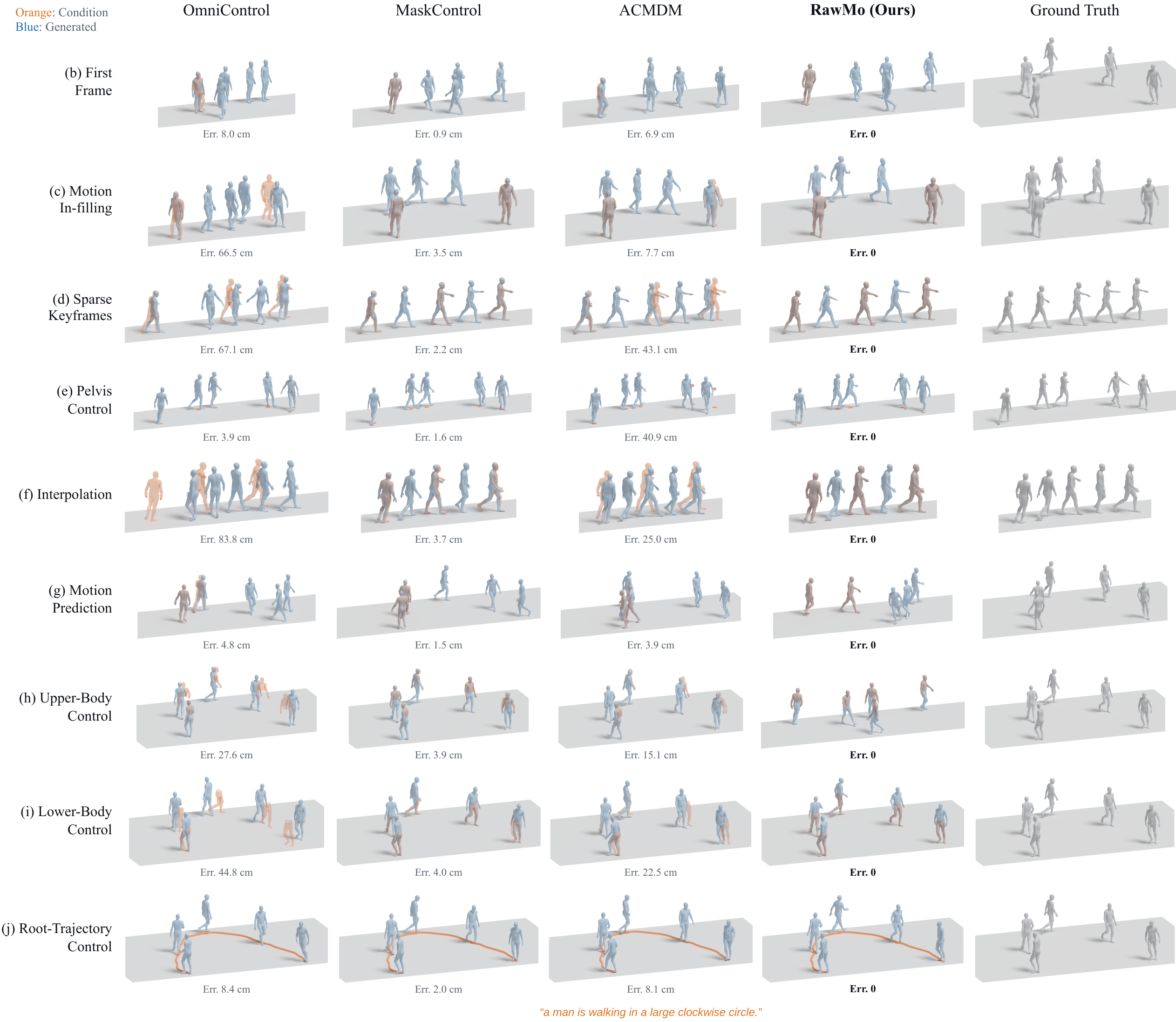}
    \caption{\textbf{Motion-conditioned generation: a clockwise walking example.} We compare RawMo with the baseline methods on a HumanML3D walking sequence under different motion conditions. The bottom row provides the horizontal root trajectory as the motion condition. RawMo records zero error at the conditioned joint positions across all nine tasks. The task order, color conventions, and error metric follow Fig.~\ref{fig:rawmo_visual_ballet}.}
    \label{fig:rawmo_visual_walk}
\end{figure*}

\noindent\textbf{Motion-conditioned generation.} Figs.~\ref{fig:rawmo_visual_ballet} and~\ref{fig:rawmo_visual_walk} extend the motion-conditioned visualizations in Fig.~\ref{fig:visual}(b)--(j) of the main paper with baseline comparisons. These comparisons also complement the quantitative evaluation in Table~\ref{tab:rawmo_conditioned_tasks} of this supplement. A single conditional RawMo model is compared with the fine-tuned OmniControl, MaskControl, and ACMDM baselines on two HumanML3D test motions, each under all nine motion-conditioning tasks. The baselines show nonzero positional errors at the conditioned joint positions, while RawMo preserves the given motion conditions exactly in every shown case.

\begin{figure*}[tp]
    \centering
    \includegraphics[width=\linewidth]{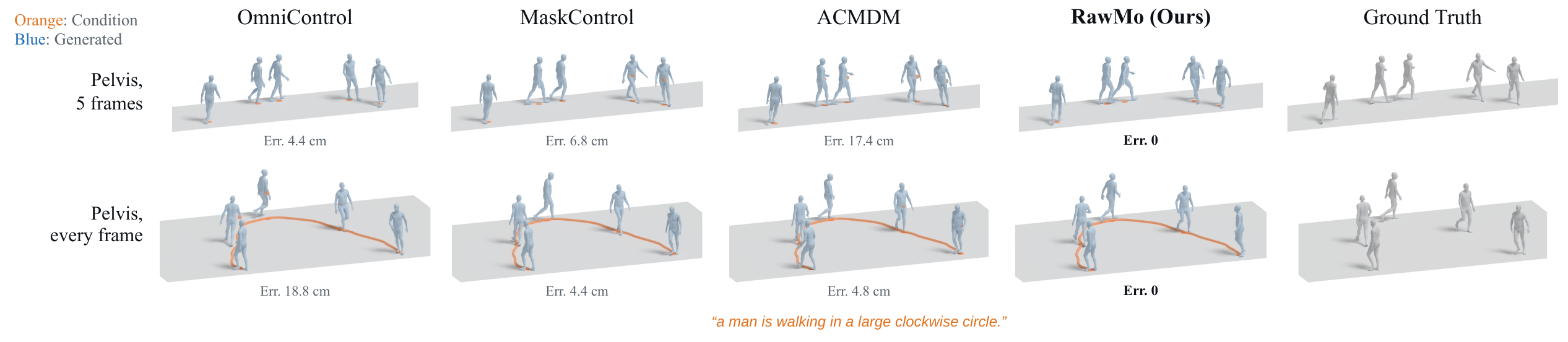}
    \caption{\textbf{Pelvis control: a clockwise walking example.} The pelvis-control visual comparisons on HumanML3D complement the quantitative results in Table~\ref{tab:cond} of the main paper. Given the pelvis positions, each method generates full-body motion using the text prompt shown in the figure. The top row provides pelvis positions at five randomly selected frames, while the bottom row provides them at every frame. Orange marks the given conditions and blue shows generated motion. RawMo preserves the conditioned pelvis positions exactly in both settings, with zero error at the controlled frames.}
    \label{fig:rawmo_pelvis_walk}
\end{figure*}

\begin{figure*}[tp]
    \centering
    \includegraphics[width=\linewidth]{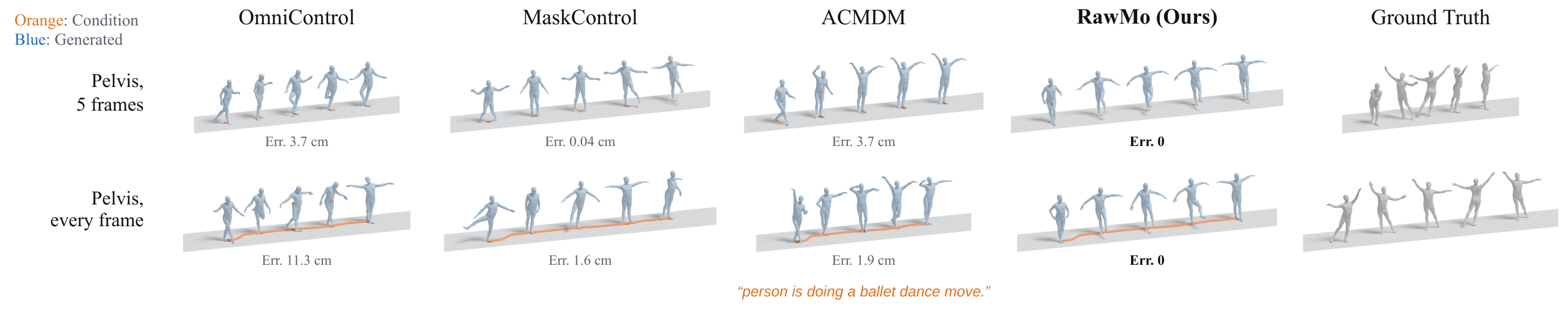}
    \caption{\textbf{Pelvis control: a ballet example.} The observation settings and visualization conventions follow Fig.~\ref{fig:rawmo_pelvis_walk}. All methods receive the given pelvis positions and the text prompt shown in the figure. The baselines show nonzero positional errors under both sparse and dense observations, while RawMo preserves the conditioned pelvis positions exactly.}
    \label{fig:rawmo_pelvis_ballet}
\end{figure*}

\noindent\textbf{Pelvis control.} Figs.~\ref{fig:rawmo_pelvis_walk} and~\ref{fig:rawmo_pelvis_ballet} provide visual comparisons corresponding to the quantitative pelvis-control results in Table~\ref{tab:cond} of the main paper. We compare sparse pelvis conditions at five randomly selected frames with a fully specified pelvis trajectory given at every frame. RawMo records zero error at the conditioned pelvis positions for both motions under both settings. In the ballet example, RawMo generates arm poses that differ from the reference while keeping the given pelvis positions unchanged.

\section{Implementation Details}
\label{supp:implementation}

\noindent\textbf{Attention contribution measurement.} Fig.~\ref{fig:analysis} in the main paper reports the relative magnitudes of attention updates contributed by frame, region, and text tokens. For each frame or region query, let $\mathcal{S}_g$ contain the key indices from source $g\in\{\mathrm{frame},\mathrm{region},\mathrm{text}\}$. Within each attention head, we compute $\sum_{j\in\mathcal{S}_g}a_j\mathbf{z}_j$, where $a_j$ is the attention weight from the query to key $j$ and $\mathbf{z}_j$ is the corresponding value vector. After concatenating heads, we apply the attention gate, output projection without its bias, root mean square layer normalization (RMSNorm)~\cite{zhang2019rmsnorm} scaling, and residual gate. We denote the resulting attention update from source $g$ in block $\ell$ by $\mathbf{u}_g^\ell$. All sources share the RMSNorm scaling computed from the full projected output, including the projection bias. The separate bias contribution is excluded from the source shares.

Let $Q\in\{\mathrm{frame},\mathrm{region}\}$ denote the query type. We use $\mathrm{E}_Q$ to first average update norms over queries of type $Q$ within each motion, then average the resulting motion-level means across motions. The contribution share of source $g$ in block $\ell$ is
\[
s^\ell_{Q,g} =
\frac{\mathrm{E}_Q[\|\mathbf{u}_g^\ell\|_2]}
     {\sum_{g'}\mathrm{E}_Q[\|\mathbf{u}_{g'}^\ell\|_2]}.
\]
For Fig.~\ref{fig:analysis}(a), we average the mean update norms over timesteps before computing the shares. For Fig.~\ref{fig:analysis}(b,c), we compute the shares separately in each block at each timestep, then average them over the 12 backbone blocks.

We average the results over five random seeds used to train RawMo for text-to-motion generation on HumanML3D. The analysis uses exponential moving average (EMA) weights and only the text-conditioned branch. All evaluations use the same 1,024 test motions and the first text prompt for each motion. For each motion and timestep $\tau\in\{0.02,0.1,0.2,0.4,0.6,0.8,0.95\}$, the noise sample is fixed across seeds.

\noindent\textbf{Motion feature representation.} On HumanML3D and KIT-ML, RawMo directly generates the motion features introduced by Guo \textit{et al.}~\cite{guo2022humanml3d}. For a skeleton with $J$ joints, each frame contains the root angular velocity about the vertical axis, two planar root-velocity components, root height, the positions and 6-D rotations of the $J-1$ non-root joints, the velocities of all $J$ joints, and four foot-contact indicators. The non-root joint positions are expressed in the root's heading frame relative to its projection onto the ground plane. This representation has $D=12J-1$ channels per frame: 263 for HumanML3D with $J=22$ and 251 for KIT-ML with $J=21$. We standardize each channel using training-split statistics. Root heading and planar position are recovered by integrating the root velocities.

\noindent\textbf{Network architecture.} Table~\ref{tab:rawmo_training_settings} summarizes the architecture, training, and sampling settings. The backbone $\mathcal{B}_{\theta}$ uses dual-stream and single-stream blocks with an MLP ratio of 4, temporal RoPE~\cite{su2024rope}, sigmoid gating on attention outputs, and RMSNorm before and after each branch. The global conditioning vector modulates each block through adaptive normalization. Each body-region token pools three to five joints over four frames and receives a learned region embedding. The 12 regions cover the pelvis, spine, chest, head, and bilateral hips, legs, shoulders, and arms. Adjacent regions share a junction joint. Each frame token is formed by projecting all $D$ features through a rank-128 bottleneck. In each refinement block, the $J\times32$ joint features of each frame are compressed into a 384-D vector. Four-head temporal attention with RoPE mixes these vectors across frames. The block then expands the resulting vectors back into joint features and applies a joint-wise MLP. A frozen CLIP ViT-B/32~\cite{radford2021clip} encodes the text prompt. All text-to-motion results use dedicated T2M models, while all motion-conditioned results use a single conditional RawMo model. The conditional model is trained on HumanML3D without HiNCE.

\begin{table*}[t]
\centering
\caption{\textbf{RawMo architecture, training, and sampling settings.} T2M denotes text-to-motion generation, and Cond.\ denotes motion-conditioned generation. The Heun row reports the number of sampling intervals, with the corresponding number of network evaluations in parentheses. CFG denotes the classifier-free guidance scale.}
\label{tab:rawmo_training_settings}
\small
\setlength{\tabcolsep}{7pt}
\renewcommand{\arraystretch}{1.1}
\resizebox{0.7\textwidth}{!}{%
\begin{tabular}{lccc}
\toprule
\rowcolor{BlueGreen!10}& \textbf{HumanML3D T2M} & \textbf{KIT-ML T2M} & \textbf{HumanML3D Cond.} \\
\midrule
Parameters & 45.7M & 49.9M & 50.5M \\
Backbone blocks (dual + single) & $8+4$ & $8+4$ & $8+4$ \\
Width $C$ / heads & $384/6$ & $384/6$ & $384/6$ \\
Refinement-head blocks & 2 & 4 & 2 \\
Extra inputs & -- & -- & $\mathbf{x}_c$, $\mathbf{m}$, root condition \\
\midrule
HiNCE weight $\lambda$ / window & $0.03$ / $\tau\leq0.25$ & $0.02$ / $\tau\leq0.25$ & not used \\
Training steps (batch size 128) & 30k & 20k & 30k \\
AdamW $(\beta_1,\beta_2)$ & $(0.9,0.95)$ & $(0.9,0.99)$ & $(0.9,0.95)$ \\
Linear decay over final & $40\%$ & $25\%$ & $40\%$ \\
Noise-level sampling $(\mu,\sigma)$ / shift & $(0.8,1.2)/5$ & $(0.8,0.8)/3$ & $(0.8,1.2)/5$ \\
\midrule
Heun intervals (network evaluations) & 16 (31) & 50 (99) & 16 (31) \\
CFG & 3.5 & 3.5 & 3 \\
\bottomrule
\end{tabular}}
\end{table*}

\noindent\textbf{Generator training.} To sample training timesteps, we first draw an initial noise fraction $\rho$ from a logit-normal distribution. We denote the resolution-shift factor by $\kappa$ and determine the training timestep $\tau$ using the resolution shift~\cite{esser2024scaling}:
\[
  1-\tau=\frac{\kappa\rho}{1+(\kappa-1)\rho}.
\]
Table~\ref{tab:rawmo_training_settings} lists the logit-normal parameters and shift factor. During training, we replace the denominator $1-\tau$ in Eq.~\ref{eq:prelim_conversion} with $\max(1-\tau,0.05)$ when converting the clean-motion prediction to flow velocity.

We use AdamW~\cite{loshchilov2019adamw} with a learning rate of $2\times10^{-4}$ and weight decay 0.01. The learning rate is warmed up over 2,000 steps, held constant, and finally decayed linearly to zero. Training uses bfloat16 precision and gradient-norm clipping at 1.0. Text prompts are dropped with probability 0.1 for classifier-free guidance. Sampling uses an exponential moving average of the generator weights with decay 0.9999.

\noindent\textbf{HiNCE implementation.} HiNCE uses a motion encoder and a text encoder. The motion encoder projects standardized motion features to 256 dimensions, reduces temporal resolution by a factor of four using two strided 1-D convolutions, and applies a four-layer, four-head Transformer. The text encoder applies a two-layer Transformer to frozen CLIP token features. Both encoders use mean pooling and output unit-normalized 512-D embeddings.

We jointly train the encoders with symmetric InfoNCE~\cite{vandenoord2018infonce} and a learned temperature initialized to 0.07. Training uses a batch size of 256 and a learning rate of $10^{-4}$, with 1,000 warm-up steps followed by cosine decay. The training budget is 12k steps on HumanML3D and 6k on KIT-ML. Checkpoints are selected by validation R@3, choosing step 3k for HumanML3D and step 1k for KIT-ML. Following TMR~\cite{petrovich2023tmr}, we mask near-duplicate negatives when the cosine similarity between the CLIP end-of-text features of the negative and positive captions exceeds 0.9. During text-to-motion generator training, both encoders remain frozen. The HiNCE loss is applied at high-noise timesteps $\tau\leq0.25$, excluding queries with dropped text prompts and masking the same near-duplicate negatives. To account for the motion encoder's temporal stride, we average the losses for the four input offsets $0,1,2,3$.

\noindent\textbf{Motion-conditioning inputs.} The conditional model receives the given motion features $\mathbf{x}_c$, their feature mask $\mathbf{m}$, and six root-conditioning channels. The six root-conditioning channels comprise two position channels, two heading channels, and two availability masks. The position channels encode the root's horizontal position in global coordinates as $(x,z)/(2\,\mathrm{m})$. The heading channels encode the root heading angle $\theta$ as $(\cos\theta,\sin\theta)$. One availability mask indicates whether the root position is given, and the other indicates whether the root heading is given. Zero-initialized linear projections inject these conditions into the joint features used for region pooling, the frame tokens, and the raw joint tokens. 

\noindent\textbf{Conditional training tasks.} The HumanML3D conditional model is trained on a mixture of text-only and motion-conditioned generation tasks. The temporal training tasks correspond to Fig.~\ref{fig:visual}(b,d,f,g) of the main paper: (b) the first frame; (d) periodic keyframes with a stride of 2--32 frames or random keyframes with a count selected from $\{1,2,3,5,8,12,20,30\}$; (f) both endpoint frames; and (g) a prefix covering the first 10--50\% of the clip. Spatial conditioning covers (h) upper-body control, (i) lower-body control, (j) root-path control, and (e) pelvis control. For pelvis control, we specify positions at 1, 2, or 5 frames, or at 10\%, 25\%, 50\%, or 100\% of the frames. Motion in-filling in Fig.~\ref{fig:visual}(c) is evaluated zero-shot.

\noindent\textbf{Preserving pelvis positions.} Pelvis control specifies the pelvis position at selected frames. At each network evaluation, the global skip copies the given pelvis height into the predicted clean motion. Horizontal pelvis positions are recovered by converting the planar root velocities to global coordinates and integrating them over motion frames. We denote the corrected horizontal displacement from frame $t$ to $t+1$ in global coordinates by $\Delta\widetilde{\mathbf{p}}_t$. The $i$-th controlled frame in temporal order has frame index $k_i$ and given horizontal position $\mathbf{p}_{k_i}$. For consecutive controlled frames $k_i<k_{i+1}$, the displacements must satisfy
\[
  \sum_{t=k_i}^{k_{i+1}-1}\Delta\widetilde{\mathbf{p}}_t
  =\mathbf{p}_{k_{i+1}}-\mathbf{p}_{k_i}.
\]
We adjust the planar root-velocity entries between these frames that are not fixed by the feature mask $\mathbf{m}$, so that the accumulated displacement satisfies this constraint. The corrected clean prediction is then converted into a flow velocity using Eq.~\ref{eq:prelim_conversion} of the main paper. The sampler uses this velocity to update the noisy motion at the next sampling step.

\noindent\textbf{Sampling.} We apply classifier-free guidance to the flow velocity and integrate from noise at $\tau=0$ to clean motion at $\tau=1$. The sampling grid is obtained by applying the same resolution shift used for training to uniformly spaced noise fractions from 1 to 0.001. Across $N$ sampling intervals, Heun's method~\cite{karras2022heun} followed by a final Euler step requires $2N-1$ network evaluations. Text-to-motion guidance scales are selected on the validation split.


\section{Limitations}
\label{supp:limitations}

\noindent\textbf{Zero-shot motion control.} RawMo performs motion control using a conditional model trained separately from the dedicated text-to-motion model. Enabling the text-to-motion model to follow motion conditions without additional training remains an open problem.

\noindent\textbf{Sampling latency.} RawMo generates motion through iterative denoising. As reported in Table~\ref{tab:cost}, generating a single HumanML3D clip on an NVIDIA RTX 4090 takes a median of 0.594\,s. Future work could focus on reducing the number of sampling steps while preserving motion quality, with the goal of lowering generation latency and improving responsiveness in interactive applications.

%% file: main.bib
@inproceedings{guo2022humanml3d,
  title={Generating diverse and natural 3d human motions from text},
  author={Guo, Chuan and Zou, Shihao and Zuo, Xinxin and Wang, Sen and Ji, Wei and Li, Xingyu and Cheng, Li},
  booktitle={2022 IEEE/CVF Conference on Computer Vision and Pattern Recognition (CVPR)},
  pages={5142--5151},
  year={2022},
  organization={IEEE}
}

@article{plappert2016kitml,
  title={The kit motion-language dataset},
  author={Plappert, Matthias and Mandery, Christian and Asfour, Tamim},
  journal={Big data},
  volume={4},
  number={4},
  pages={236--252},
  year={2016},
  publisher={SAGE Publications, Sage CA: Los Angeles, CA}
}

@inproceedings{tevet2023mdm,
title={Human Motion Diffusion Model},
author={Guy Tevet and Sigal Raab and Brian Gordon and Yoni Shafir and Daniel Cohen-or and Amit Haim Bermano},
booktitle={The Eleventh International Conference on Learning Representations },
year={2023},
url={https://openreview.net/forum?id=SJ1kSyO2jwu}
}

@inproceedings{chen2023mld,
  title={Executing your commands via motion diffusion in latent space},
  author={Chen, Xin and Jiang, Biao and Liu, Wen and Huang, Zilong and Fu, Bin and Chen, Tao and Yu, Gang},
  booktitle={2023 IEEE/CVF Conference on Computer Vision and Pattern Recognition (CVPR)},
  pages={18000--18010},
  year={2023},
  organization={IEEE}
}

@inproceedings{zhang2023t2mgpt,
  title={Generating human motion from textual descriptions with discrete representations},
  author={Zhang, Jianrong and Zhang, Yangsong and Cun, Xiaodong and Zhang, Yong and Zhao, Hongwei and Lu, Hongtao and Shen, Xi and Shan, Ying},
  booktitle={Proceedings of the IEEE/CVF conference on computer vision and pattern recognition},
  pages={14730--14740},
  year={2023}
}

@inproceedings{guo2024momask,
  title={Momask: Generative masked modeling of 3d human motions},
  author={Guo, Chuan and Mu, Yuxuan and Javed, Muhammad Gohar and Wang, Sen and Cheng, Li},
  booktitle={2024 IEEE/CVF Conference on Computer Vision and Pattern Recognition (CVPR)},
  pages={1900--1910},
  year={2024},
  organization={IEEE}
}

@inproceedings{li2026jit,
  title={Back to basics: Let denoising generative models denoise},
  author={Li, Tianhong and He, Kaiming},
  booktitle={Proceedings of the IEEE/CVF Conference on Computer Vision and Pattern Recognition},
  pages={36115--36125},
  year={2026}
}

@inproceedings{yu2026pixeldit,
  title={Pixeldit: Pixel diffusion transformers for image generation},
  author={Yu, Yongsheng and Xiong, Wei and Nie, Weili and Sheng, Yichen and Liu, Shiqiu and Luo, Jiebo},
  booktitle={Proceedings of the IEEE/CVF Conference on Computer Vision and Pattern Recognition},
  pages={14273--14282},
  year={2026}
}

@article{ho2020ddpm,
  title={Denoising diffusion probabilistic models},
  author={Ho, Jonathan and Jain, Ajay and Abbeel, Pieter},
  journal={Advances in neural information processing systems},
  volume={33},
  pages={6840--6851},
  year={2020}
}

@article{dhariwal2021adm,
  title={Diffusion models beat gans on image synthesis},
  author={Dhariwal, Prafulla and Nichol, Alexander},
  journal={Advances in neural information processing systems},
  volume={34},
  pages={8780--8794},
  year={2021}
}

@inproceedings{rombach2022ldm,
  title={High-resolution image synthesis with latent diffusion models},
  author={Rombach, Robin and Blattmann, Andreas and Lorenz, Dominik and Esser, Patrick and Ommer, Bj{\"o}rn},
  booktitle={2022 IEEE/CVF conference on computer vision and pattern recognition (CVPR)},
  pages={10674--10685},
  year={2022},
  organization={ieee}
}

@inproceedings{chen2026dip,
  title={Dip: Taming diffusion models in pixel space},
  author={Chen, Zhennan and Zhu, Junwei and Chen, Xu and Zhang, Jiangning and Hu, Xiaobin and Zhao, Hanzhen and Wang, Chengjie and Yang, Jian and Tai, Ying},
  booktitle={Proceedings of the IEEE/CVF Conference on Computer Vision and Pattern Recognition},
  pages={36136--36146},
  year={2026}
}

@article{zhang2024motiondiffuse,
  title={Motiondiffuse: Text-driven human motion generation with diffusion model},
  author={Zhang, Mingyuan and Cai, Zhongang and Pan, Liang and Hong, Fangzhou and Guo, Xinying and Yang, Lei and Liu, Ziwei},
  journal={IEEE transactions on pattern analysis and machine intelligence},
  volume={46},
  number={6},
  pages={4115--4128},
  year={2024},
  publisher={IEEE}
}

@inproceedings{zhang2023remodiffuse,
  title={Remodiffuse: Retrieval-augmented motion diffusion model},
  author={Zhang, Mingyuan and Guo, Xinying and Pan, Liang and Cai, Zhongang and Hong, Fangzhou and Li, Huirong and Yang, Lei and Liu, Ziwei},
  booktitle={2023 IEEE/CVF International Conference on Computer Vision (ICCV)},
  pages={364--373},
  year={2023},
  organization={IEEE}
}

@inproceedings{li2026motionhiflow,
  title     = {MotionHiFlow: Text-to-Motion via Hierarchical Flow Matching},
  author    = {Li, Heng and Lin, Xiaotong and Zeng, Ling-An and Kang, Yulei and Li, Shuai and Hu, Jian-Fang},
  booktitle = {Proceedings of the IEEE/CVF Conference on Computer Vision and Pattern Recognition (CVPR)},
  year      = {2026}
}

@article{yuan2024mogents,
  title={Mogents: Motion generation based on spatial-temporal joint modeling},
  author={Yuan, Weihao and He, Yisheng and Shen, Weichao and Dong, Yuan and Gu, Xiaodong and Dong, Zilong and Bo, Liefeng and Huang, Qixing},
  journal={Advances in Neural Information Processing Systems},
  volume={37},
  pages={130739--130763},
  year={2024}
}

@inproceedings{hong2025salad,
  title={Salad: Skeleton-aware latent diffusion for text-driven motion generation and editing},
  author={Hong, Seokhyeon and Kim, Chaelin and Yoon, Serin and Nam, Junghyun and Cha, Sihun and Noh, Junyong},
  booktitle={2025 IEEE/CVF Conference on Computer Vision and Pattern Recognition (CVPR)},
  pages={7158--7168},
  year={2025},
  organization={IEEE}
}

@article{yu2026motionspaceflow,
  title={MotionSpaceFlow: Representation-Aware Flow Matching in Direct Motion Space},
  author={Yu, Qing and Fujiwara, Kent},
  journal={arXiv preprint arXiv:2609.34190},
  year={2026}
}

@inproceedings{karunratanakul2023gmd,
  title={Guided motion diffusion for controllable human motion synthesis},
  author={Karunratanakul, Korrawe and Preechakul, Konpat and Suwajanakorn, Supasorn and Tang, Siyu},
  booktitle={2023 IEEE/CVF International Conference on Computer Vision (ICCV)},
  pages={2151--2162},
  year={2023},
  organization={IEEE}
}

@inproceedings{shafir2024priormdm,
  title={Human motion diffusion as a generative prior},
  author={Shafir, Yonatan and Tevet, Guy and Kapon, Roy and Bermano, Amit},
  booktitle={International Conference on Learning Representations},
  volume={2024},
  pages={8717--8733},
  year={2024}
}

@inproceedings{xie2024omnicontrol,
  title={Omnicontrol: Control any joint at any time for human motion generation},
  author={Xie, Yiming and Jampani, Varun and Zhong, Lei and Sun, Deqing and Jiang, Huaizu},
  booktitle={International Conference on Learning Representations},
  volume={2024},
  pages={28176--28194},
  year={2024}
}

@inproceedings{dai2024motionlcm,
  title={Motionlcm: Real-time controllable motion generation via latent consistency model},
  author={Dai, Wenxun and Chen, Ling-Hao and Wang, Jingbo and Liu, Jinpeng and Dai, Bo and Tang, Yansong},
  booktitle={European Conference on Computer Vision},
  pages={390--408},
  year={2024},
  organization={Springer}
}

@inproceedings{wan2024tlcontrol,
  title={Tlcontrol: Trajectory and language control for human motion synthesis},
  author={Wan, Weilin and Dou, Zhiyang and Komura, Taku and Wang, Wenping and Jayaraman, Dinesh and Liu, Lingjie},
  booktitle={European Conference on Computer Vision},
  pages={37--54},
  year={2024},
  organization={Springer}
}

@inproceedings{cohan2024condmdi,
  title={Flexible motion in-betweening with diffusion models},
  author={Cohan, Setareh and Tevet, Guy and Reda, Daniele and Peng, Xue Bin and Van de Panne, Michiel},
  booktitle={ACM SIGGRAPH 2024 conference papers},
  pages={1--9},
  year={2024}
}

@inproceedings{guo2025motionlab,
  title={Motionlab: Unified human motion generation and editing via the motion-condition-motion paradigm},
  author={Guo, Ziyan and Hu, Zeyu and Soh, De Wen and Zhao, Na},
  booktitle={Proceedings of the IEEE/CVF International Conference on Computer Vision},
  pages={13869--13879},
  year={2025}
}

@inproceedings{watanabe2026projflow,
  title   = {ProjFlow: Projection Sampling with Flow Matching for Zero-Shot Exact Spatial Motion Control},
  author  = {Akihisa Watanabe and Qing Yu and Edgar Simo-Serra and Kent Fujiwara},
  booktitle={Proceedings of the IEEE/CVF Conference on Computer Vision and Pattern Recognition},
  year    = {2026}
}

@inproceedings{yu2026cmdm,
      title={Causal Motion Diffusion Models for Autoregressive Motion Generation},
      author={Yu, Qing and Watanabe, Akihisa and Fujiwara, Kent},
      booktitle={CVPR},
      year={2026}
  }

@article{radford2021clip,
  title={Learning transferable visual models from natural language supervision},
  author={Radford, A},
  journal={arXiv preprint arXiv:2103.00020},
  year={2021}
}

@misc{blackforestlabs2024flux,
    author={Black Forest Labs},
    title={FLUX},
    year={2024},
    howpublished={\url{https://github.com/black-forest-labs/flux}},
}

@article{su2024rope,
  title={Roformer: Enhanced transformer with rotary position embedding},
  author={Su, Jianlin and Ahmed, Murtadha and Lu, Yu and Pan, Shengfeng and Bo, Wen and Liu, Yunfeng},
  journal={Neurocomputing},
  volume={568},
  pages={127063},
  year={2024},
  publisher={Elsevier}
}

@article{vandenoord2018infonce,
  title={Representation learning with contrastive predictive coding},
  author={Oord, Aaron van den and Li, Yazhe and Vinyals, Oriol},
  journal={arXiv preprint arXiv:1807.03748},
  year={2018}
}

@inproceedings{meng2025mardm,
  title={Rethinking diffusion for text-driven human motion generation: Redundant representations, evaluation, and masked autoregression},
  author={Meng, Zichong and Xie, Yiming and Peng, Xiaogang and Han, Zeyu and Jiang, Huaizu},
  booktitle={2025 IEEE/CVF Conference on Computer Vision and Pattern Recognition (CVPR)},
  pages={27859--27871},
  year={2025},
  organization={IEEE}
}

@article{karras2022heun,
  title={Elucidating the design space of diffusion-based generative models},
  author={Karras, Tero and Aittala, Miika and Aila, Timo and Laine, Samuli},
  journal={Advances in neural information processing systems},
  volume={35},
  pages={26565--26577},
  year={2022}
}

@inproceedings{pinyoanuntapong2024mmm,
  title={Mmm: Generative masked motion model},
  author={Pinyoanuntapong, Ekkasit and Wang, Pu and Lee, Minwoo and Chen, Chen},
  booktitle={2024 IEEE/CVF Conference on Computer Vision and Pattern Recognition (CVPR)},
  pages={1546--1555},
  year={2024},
  organization={IEEE}
}

@inproceedings{pinyoanuntapong2024bamm,
  title={Bamm: Bidirectional autoregressive motion model},
  author={Pinyoanuntapong, Ekkasit and Saleem, Muhammad Usama and Wang, Pu and Lee, Minwoo and Das, Srijan and Chen, Chen},
  booktitle={European Conference on Computer Vision},
  pages={172--190},
  year={2024},
  organization={Springer}
}

@inproceedings{zeng2025lightt2m,
  title={Light-t2m: A lightweight and fast model for text-to-motion generation},
  author={Zeng, Ling-An and Huang, Guohong and Wu, Gaojie and Zheng, Wei-Shi},
  booktitle={Proceedings of the AAAI Conference on Artificial Intelligence},
  volume={39},
  number={9},
  pages={9797--9805},
  year={2025}
}

@inproceedings{zhang2025momadiff,
  title={Towards robust and controllable text-to-motion via masked autoregressive diffusion},
  author={Zhang, Zongye and Kong, Bohan and Liu, Qingjie and Wang, Yunhong},
  booktitle={Proceedings of the 33rd ACM International Conference on Multimedia},
  pages={9326--9335},
  year={2025}
}

@inproceedings{hong2026bipo,
  title={Bipo: Bidirectional partial occlusion network for text-to-motion synthesis},
  author={Hong, Seong-Eun and Lim, Soobin and Hwang, Juyeong and Chang, Minwook and Kang, Hyeongyeop},
  booktitle={2026 IEEE/CVF Winter Conference on Applications of Computer Vision (WACV)},
  pages={32--42},
  year={2026},
  organization={IEEE}
}

@article{wang2024intercontrol,
  title={Intercontrol: Zero-shot human interaction generation by controlling every joint},
  author={Wang, Zhenzhi and Wang, Jingbo and Li, Yixuan and Lin, Dahua and Dai, Bo},
  journal={Advances in Neural Information Processing Systems},
  volume={37},
  pages={105397--105424},
  year={2024}
}

@inproceedings{pinyoanuntapong2025maskcontrol,
  title={Maskcontrol: Spatio-temporal control for masked motion synthesis},
  author={Pinyoanuntapong, Ekkasit and Saleem, Muhammad and Karunratanakul, Korrawe and Wang, Pu and Xue, Hongfei and Chen, Chen and Guo, Chuan and Cao, Junli and Ren, Jian and Tulyakov, Sergey},
  booktitle={Proceedings of the IEEE/CVF International Conference on Computer Vision},
  pages={9955--9965},
  year={2025}
}

@article{meng2025acmdm,
  title={Absolute coordinates make motion generation easy},
  author={Meng, Zichong and Han, Zeyu and Peng, Xiaogang and Xie, Yiming and Jiang, Huaizu},
  journal={arXiv preprint arXiv:2505.19377},
  year={2025}
}

@inproceedings{fan2025motionmillion,
  title={Go to zero: Towards zero-shot motion generation with million-scale data},
  author={Fan, Ke and Lu, Shunlin and Dai, Minyue and Yu, Runyi and Xiao, Lixing and Dou, Zhiyang and Dong, Junting and Ma, Lizhuang and Wang, Jingbo},
  booktitle={2025 IEEE/CVF International Conference on Computer Vision (ICCV)},
  pages={13336--13348},
  year={2025},
  organization={IEEE}
}

@inproceedings{zhang2026romo,
  title={RoMo: A large-scale, richly organized dataset and semantic taxonomy for human motion generation},
  author={Zhang, Jiahao and Liu, Joseph and Lee, Young-Yoon and Moon, Seonghyeon and Zordan, Victor and Tevet, Guy and Liu, C Karen and Gould, Stephen and Jacob, Oren and Jiang, Haomiao and others},
  booktitle={Proceedings of the IEEE/CVF Conference on Computer Vision and Pattern Recognition},
  pages={16408--16419},
  year={2026}
}

@inproceedings{xiao2025motionstreamer,
  title={Motionstreamer: Streaming motion generation via diffusion-based autoregressive model in causal latent space},
  author={Xiao, Lixing and Lu, Shunlin and Pi, Huaijin and Fan, Ke and Pan, Liang and Zhou, Yueer and Feng, Ziyong and Zhou, Xiaowei and Peng, Sida and Wang, Jingbo},
  booktitle={2025 IEEE/CVF International Conference on Computer Vision (ICCV)},
  pages={10086--10096},
  year={2025},
  organization={IEEE}
}

@article{loshchilov2019adamw,
  title={Decoupled weight decay regularization},
  author={Loshchilov, Ilya and Hutter, Frank},
  journal={arXiv preprint arXiv:1711.05101},
  year={2017}
}

@inproceedings{esser2024scaling,
  title={Scaling rectified flow transformers for high-resolution image synthesis},
  author={Esser, Patrick and Kulal, Sumith and Blattmann, Andreas and Entezari, Rahim and M{\"u}ller, Jonas and Saini, Harry and Levi, Yam and Lorenz, Dominik and Sauer, Axel and Boesel, Frederic and others},
  booktitle={Forty-first international conference on machine learning},
  year={2024}
}

@inproceedings{petrovich2023tmr,
  title={TMR: Text-to-motion retrieval using contrastive 3D human motion synthesis},
  author={Petrovich, Mathis and Black, Michael J and Varol, G{\"u}l},
  booktitle={2023 IEEE/CVF International Conference on Computer Vision (ICCV)},
  pages={9454--9463},
  year={2023},
  organization={IEEE}
}

@article{zhang2019rmsnorm,
  title={Root mean square layer normalization},
  author={Zhang, Biao and Sennrich, Rico},
  journal={Advances in neural information processing systems},
  volume={32},
  year={2019}
}
